\documentclass[sigconf]{acmart} 
\AtBeginDocument{%
  \providecommand\BibTeX{{%
    \normalfont B\kern-0.5em{\scshape i\kern-0.25em b}\kern-0.8em\TeX}}}

\setcopyright{none}
\renewcommand\footnotetextcopyrightpermission[1]{}
\usepackage{algorithm}
\usepackage{algorithmic}
\usepackage{indentfirst}
\usepackage{CJK}

\usepackage{multirow}
\usepackage{amsmath, amsthm, bbm, mathtools, commath}
\usepackage{datetime}
\usepackage{subcaption}
\usepackage{colortbl}
\usepackage{arydshln} %

\newcommand{\printfnsymbol}[1]{%
  \textsuperscript{\@fnsymbol{#1}}%
}
\makeatother

\newcommand{\std}[1]{\pm #1}

\newcommand{\blue}[1]{{\color{blue}#1}}
\graphicspath{{photo/}}
\begin{document}

\title{Unifying Adversarial Perturbation for Graph Neural Networks}


\author{Jinluan Yang\textsuperscript{*}, Ruihao Zhang\textsuperscript{*}, Zhengyu Chen, Fei Wu, Kun Kuang
\\
 Zhejiang University
\\
\texttt{yangjinluan@zju.edu.cn}
\footnotetext{ \textsuperscript{*}These authors contributed equally to this work.}
}

% \author{Anonymous Authors}
% \affiliation{%

% \affiliation{%

% \affiliation{%

%% other information printed in the page headers. This command allows .
\renewcommand{\shortauthors}{Yang et al. }

\begin{abstract}
This paper studies the vulnerability of Graph Neural Networks (GNNs) to adversarial attacks on node features and graph structure. 
Various methods have implemented adversarial training to augment graph data, aiming to bolster the robustness and generalization of GNNs. These methods typically involve applying perturbations to the node feature, weights, or graph structure and subsequently minimizing the loss by learning more robust graph model parameters under the adversarial perturbations.
Despite the effectiveness of adversarial training in enhancing GNNs' robustness and generalization abilities, its application has been largely confined to specific datasets and GNN types. In this paper, we propose a novel method, PerturbEmbedding, that integrates adversarial perturbation and training, enhancing GNNs' resilience to such attacks and improving their generalization ability. PerturbEmbedding performs perturbation operations directly on every hidden embedding of GNNs and provides a unified framework for most existing perturbation strategies/methods. We also offer a unified perspective on the forms of perturbations, namely random and adversarial perturbations. 
Through experiments on various datasets using different backbone models, we demonstrate that PerturbEmbedding significantly improves both the robustness and generalization abilities of GNNs, outperforming existing methods. The rejection of both random (non-targeted) and adversarial (targeted) perturbations further enhances the backbone model's performance.

\end{abstract}






\maketitle

\section{Introduction}
\begin{figure*}[]
\centering 
\includegraphics[width=14cm]{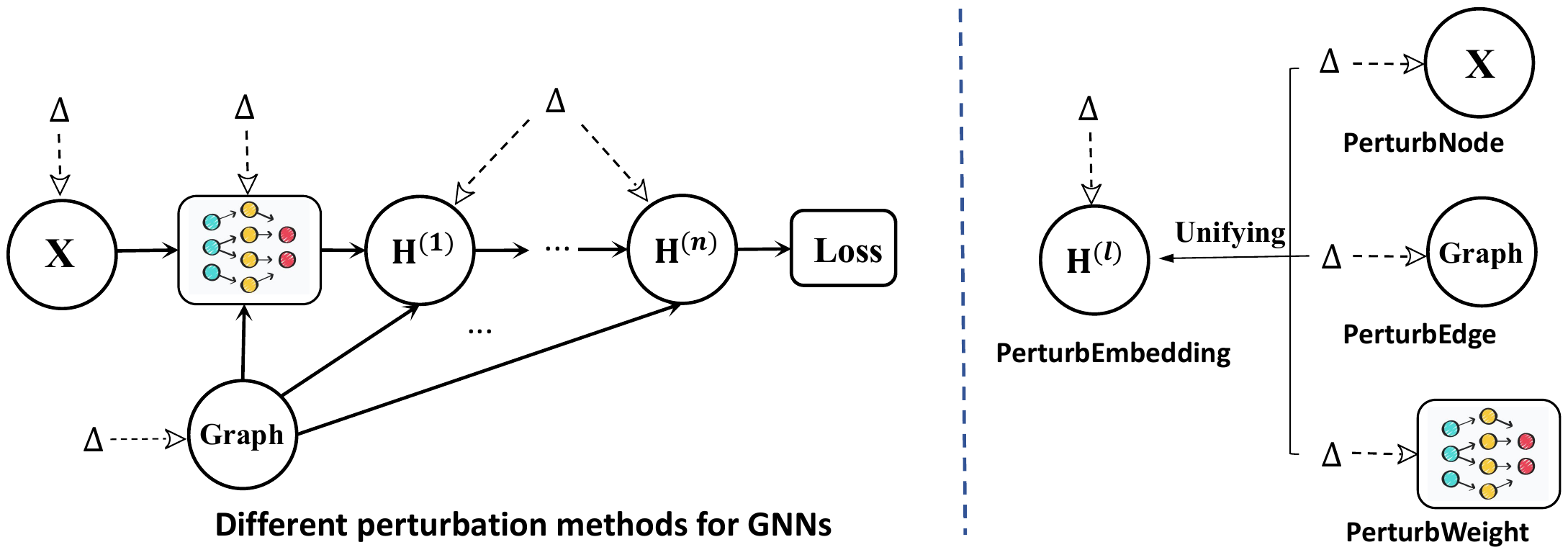}  % 图片路径和大小
\caption{Illustrations of PerturbEmbedding and other existing perturbation methods. Existing perturbation methods mainly apply perturbations to
the node feature (PerturbNode), weights (PerturbWeight) or graph structure (PerturbEdge), while PerturbEmbedding performs perturbation operations directly on every hidden embedding of GNNs.  %IHGL contains two modules, the environment clustering module aims to learn multiple graph partitions, and the invariant graph learning module aims to learn the invariant representation based on multiple graph partitions.
}    % 整个大图的标题
\label{fig:framework}       % 标签
\end{figure*}

%\blue{
Graph structured data is ubiquitous in many real-world applications, such as social networks \cite{tang2009social,peng2022reinforced}, knowledge bases \cite{vashishth2019composition} and recommendation systems \cite{chen2021deep, chen2019deep}.  As powerful tools for representation learning on graphs, graph neural networks (GNNs) have shown promising results in many graph-related tasks such as node classification \cite{chen2022ba} and graph classification \cite{defferrard2016convolutional}. However, despite the great success of Graph neural networks, recent studies have shown that GNNs are vulnerable to adversarial attacks on node features and graph structure \cite{deng2019batch, xu2019topology, athalye2018obfuscated}. Moreover, Due to the additional consideration of connections between examples, GNNs could be more sensitive to the perturbations since the perturbations from neighbor nodes exacerbate the impact on a target node \cite{feng2019graph}.%}

%\blue{
The sensitivity of GNNs to perturbations, particularly those that propagate through the graph's structure, underscores the need for robust training strategies. Adversarial training is one strategy that has been proposed to mitigate this sensitivity \cite{madry2017towards, goodfellow2014explaining, shafahi2019adversarial, athalye2018obfuscated, miyato2016adversarial}.
%Among the recent emergence of promising techniques, adversarial training \cite{madry2017towards, goodfellow2014explaining, shafahi2019adversarial, athalye2018obfuscated, miyato2016adversarial} has been empirically demonstrated to achieve the superiority in terms of tackling the vulnerability to intentional perturbations of deep neural networks on image and text data. 
The key to adversarial training is augmenting data with the worst-case adversarial examples, such that the trained model can
correctly classify future adversarial examples. Recently methods \cite{feng2019graph, xu2019topology, zhu2019robust, zugner2019certifiable, jin2020graph, kong2020flag, wu2020adversarial} have implemented adversarial training to augment graph data, aiming to bolster the robustness and generalization of GNNs. These methods typically involve applying perturbations to the node feature, weights, or graph structure and subsequently minimizing the loss by learning more robust graph model parameters under the perturbations. {Specifically, perturbations can be further categorized into targeted and non-targeted based on whether they are generated using specific label information. Random perturbations are non-targeted and do not rely on specific label information, while adversarial perturbations are targeted, intentional, and take the target label into account when calculating the loss.} Despite the effectiveness of adversarial training in enhancing GNNs' robustness and generalization abilities, its application has been largely confined to specific datasets and GNN types. %The absence of a theoretical explanation for the effectiveness of perturbation methods on GNNs further complicates the matter.

% Despite its success in the image domain, adversarial training does not (yet) stand out as an effective defense for Graph Neural Networks (GNNs) against graph structure perturbations.

This paper studies the vulnerability of Graph Neural Networks (GNNs) to adversarial attacks on node features and graph structure. Several open problems related to adversarial training on GNNs persist. Firstly, finding a universal perturbation strategy/method suitable for all cases, considering the divergence of different datasets and models is a daunting task. Secondly, the generation of adversarial perturbations in previous methods primarily relied on gradient information, which is a time-consuming process. Thus, discovering a simple and efficient method still remains a challenge.

We propose a novel method, PerturbEmbedding, that integrates adversarial perturbation and training, enhancing GNNs' resilience to such attacks and improving their generalization ability. PerturbEmbedding performs perturbation operations directly on every hidden embedding of GNNs and provides a unified framework for most existing perturbation strategies/methods. We also offer a unified perspective on the forms of perturbations, namely random (non-targeted) and adversarial (targeted) perturbations. Through experiments on various datasets using different backbone models, we demonstrate that PerturbEmbedding significantly improves both the robustness and generalization abilities of GNNs. The rejection of both random (non-targeted) and adversarial (targeted) perturbations further enhances the backbone model's performance.  Our contributions can be concluded as follows:
\begin{itemize}
    \item We propose a novel method, PerturbEmbedding, which performs perturbation operations directly on every hidden embedding of GNNs and provides a unified framework for most existing perturbation strategies/methods. In other words, these perturbation strategies/methods can be regarded as one special form of PerturbEmbedding.
    \item We offer a unified perspective on the forms of perturbations, namely random (non-targeted) and adversarial perturbations (targeted), to understand the mechanism of perturbation.
    \item We conduct experiments on various datasets using different backbone models to demonstrate that PerturbEmbedding significantly improves both the robustness and generalization abilities of GNNs, outperforming existing methods.
    
\end{itemize}

\section{Related work}
\subsection{Graph Neural Networks}
Graph Neural Networks (GNNs) have shown promising results in various graph-based applications \cite{chen2024learning,lv2025debiased,lv2025grasp,lv2025out}. Based on the strong homophily assumption that nodes with similar properties are more likely to be linked together, \cite{bruna2013spectral} firstly propose a Convolutional Neural Network in graph domain based on the spectrum of the graph Laplacian. GCN \cite{kipf2016semi} introduces a simple and well-behaved layer-wise propagation rule for neural network models that operate directly on graphs 
% and show how it can be motivated from a first-order approximation of spectral graph convolutions.
GAT \cite{velickovic2017graph} introduces an attention-based architecture, the idea is to compute the hidden representations of each node leveraging masked self-attentional layers. However, such an assumption of homophily is not always true for heterophilic graphs \cite{zhang2024discovering,yang2025leveraging}, for example, in online transaction networks, fraudsters are more likely to build connections with customers instead of other fraudsters. To make GNNs generalize to heterophilic graphs, MixHop \cite{abu2019mixhop} Proposes a mixed feature model that aggregates messages from multi-hop neighbors by mixing powers of the adjacency matrix. 
% H2GCN \cite{zhu2020beyond} applies three useful designs—ego- and neighbor-embedding separation, higher-order neighborhoods, and a combination of intermediate representations that boost learning from the graph structure under low-homophily settings. 
LINKX \cite{lim2021large} proposes a simple method that combines two simple baselines MLP and LINK, which overcomes the scalable issues on large graphs. 

%\blue{However, recent works \cite{bojchevski2019adversarial, dai2018adversarial} have demonstrated that GNNs are vulnerable to small but intentional perturbations on the 1) node features and/or 2) the structure of the graph (i.e., adding and/or deleting edges in the graph). Therefore, it is of great importance to develop robust GNNs against such attacks.}

Despite these advancements, recent studies \cite{bojchevski2019adversarial, dai2018adversarial} have revealed that GNNs are susceptible to adversarial attacks, which involve small but intentional perturbations to node features or graph structure, such as adding or deleting edges. This vulnerability underscores the critical need for developing robust GNNs capable of withstanding such targeted manipulations.

% However, recent works \cite{bojchevski2019adversarial, dai2018adversarial} have demonstrated that an attacker can easily fool GNNs to make incorrect predictions via perturbing 1) the node features and/or 2) the structure of the graph (i.e., adding and/or deleting edges in the graph)

\subsection{Adversarial Training}
Adversarial training \cite{madry2017towards, goodfellow2014explaining, shafahi2019adversarial, athalye2018obfuscated, miyato2016adversarial,yang2024mitigating,liu2024mmi,liu2025adversarial} is a widely used countermeasure to tackle the vulnerability to intentional perturbations of deep neural networks on image and text data, which involves an alternative min-max process. The key of adversarial training is generating adversarial examples from clean examples with perturbations maximally attacking the training objective such that the trained model can correctly classify the future adversarial examples. 

{Specifically, the perturbations can be further categorized into targeted and non-targeted based on whether they are generated using specific label information.} {Random perturbations are used to generate adversarial samples, which aim to slightly modify the sample in such a way that the target model misclassifies it without any particular preference towards a specific output. These perturbations are small, random changes added to the input data with the intent to deceive the model into making an incorrect classification decision. Since random perturbations do not rely on specific label information, they are not designed to misclassify the input data into any particular category but rather to cause the model's prediction to be incorrect. Hence, this type of attack is considered non-targeted. Targeted attacks are designed to generate adversarial samples that can change the prediction of a model to a specific target label. In these attacks, the perturbations are purposeful and consider the target label when calculating the loss. Attackers optimize a loss function that quantifies the discrepancy between the model's output and the desired target label to generate adversarial perturbations. Therefore, these perturbations are intentional and meant to cause the model to misclassify the input data into a specific, attacker-chosen category. Since the loss calculation involves a specific label, this type of attack is considered targeted.} In mathematical terms, adversarial training is posed as a min-max problem, aiming to find the optimal solution for the worst-case scenario. \cite{madry2017towards} showed that this min-max problem could be reliably tackled by Stochastic Gradient Descent (SGD) for the outer minimization and Projected Gradient Descent (PGD) for the inner maximization. \cite{miyato2016adversarial} extends adversarial and virtual adversarial training to the text domain by applying perturbations to the word embeddings in a recurrent neural network rather than to the original input itself. \cite{shafahi2019adversarial} presents an algorithm that eliminates the overhead cost of generating adversarial examples by recycling the gradient information computed when updating model parameters.

%\blue{However, directly adopting Adversarial Training on GNNs is less effective since Adversarial Training regards examples as independent of each other and does not consider the impact from connected examples. Moreover, Due to the additional consideration of connections between examples, graph neural networks could be more sensitive to the perturbations since the perturbations from neighbor nodes exacerbate the impact on a target node.}

However, adversarial training for GNNs presents challenges. GNNs inherently treat examples as interconnected due to the graph structure, unlike the independent assumption in traditional adversarial training. This interconnectedness means that GNNs are particularly sensitive to perturbations, as changes in one node can propagate through the graph and significantly affect other nodes. The adjacency matrix, which captures these connections, can amplify the impact of adversarial perturbations, making GNNs more vulnerable to attacks that exploit the graph's topology.

\subsection{Adversarial Training for Graphs}

Several studies \cite{deng2019batch, xu2019topology, athalye2018obfuscated} have shown that GNNs are vulnerable to adversarial attacks on the node features and/or graph structure. Recently many methods have applied adversarial training to augment graph data to to improve the robustness and generalization of GNNs. Specifically, these methods mainly apply perturbation to the node feature, weights or graph structure. \cite{dai2018adversarial} suggest dropping edges randomly in adversarial training to generate perturbations on the adjacency matrix A. \cite{feng2019graph} designs a dynamic regularizer forcing GNN models to learn to prevent the propagation of perturbations on graphs, enhancing the model’s robustness against perturbations on node input features. \cite{xu2019topology} introduce projection gradient descent (PGD) to optimize edge
perturbation. 
\cite{dai2019adversarial} defines adversarial perturbations in the embedding space with an adaptive $L_2 $ norm constraint that depends on the connectivity pattern of node pairs.
\cite{kong2020flag} regard adversarial training as a method of data augmentation and propose FLAG which iteratively augments node features with gradient-based adversarial perturbations during training, and boosts performance at test time. \cite{jin2019latent} raise a latent adversarial training method that injects perturbations on the hidden layer.
% \cite{xue2021cap} construct the co-adversarial perturbation optimization problem in terms of weights and features. 

In contrast to these specialized methods, our work aims to develop a unified adversarial training framework that encapsulates a diverse set of adversarial strategies. This framework is designed to comprehensively alleviate the vulnerabilities of GNNs by targeting different aspects of the graph data, namely the adjacency matrix (A), node features (X), and hidden representations (H). By integrating these strategies, our framework aims to provide a more holistic defense against adversarial attacks, ensuring that GNNs can robustly learn and generalize across a wide range of graph-based tasks.

% \blue{Graph Adv methods,
% why A,X,H ?
% adversarial training

% Different from previous works, we aim to design a unified framework, which contains different adversarial \textbf{training strategies} (). }

% \subsection{Generalized GNN}

\section{Notations}
\textbf{Notations.} Let $G = (V, E)$ be a graph, where $V$ is the set of nodes, $E$ is the set of edges. $X$ $\in$ $R^{N \times F}$ denotes feature matrix, where the vector $x_v$ corresponds to the feature of node $v$, and c is the dimensionality of features. $A \in \{0,1\}^{N \times N}$ is the adjacency matrix, where $A_{uv} = 1$ if there exists an edge between node $v_i$ and $v_j$, otherwise $A_{ij} = 0$. For semi-supervised node classification tasks, only part of nodes have known labels $Y^o = \{y_1,y_2,...y_n\}$, where $y_j \in \{0,1,...,c-1\}$ denotes the label of node $j$, $c$ is the number of classes. The goal of GNNs is to predict the class of unlabeled nodes.

\section{Method}
In this section, we introduce our proposed unified framework PerturbEmbedding, which is universally applicable to all Graph Neural Networks (GNNs). 
We begin by delineating the intricacies of our approach and subsequently demonstrate
that the most common existing perturbing methods, namely PerturbNode, PerturbEdge, and PerturbWeight, can be unified into our framework PerturbEmbedding. 
After that, we offer a unified perspective on
the forms of perturbations, namely random (non-targeted) and adversarial perturbations (targeted).

\subsection{A Unified Specific framework}
As Figure \ref{fig:framework} suggests, most existing adversarial GNNs mainly apply perturbations to the node features, weights or graph structure, and then minimize the loss by learning the more robust graph model parameters $\theta$ under the perturbed samples. For simplicity, we use \emph{PerturbEdge}, \emph{PerturbNode} and \emph{PerturbWeight} to refer to these perturbation strategies in the following text. We prove that the effects of the most common existing perturbation strategies, i.e., PerturbEdge, PerturebNode and PerturbWeight can be unified into our proposed framework--directly perturbing the embedding process, which is referred to \emph{PerturbEmbedding}. 

\textbf{Unifying perturbation Strategies}. As shown in Figure \ref{fig:framework}, in intuition, applying perturbation to features, edges or weights will all ultimately act on the embedding $\mathbf{H}$. This inspires us to explore the specific connection between different perturbation strategies. As a starting point, we demonstrate that PerturbEdge, PerturbNode, PerturbWeight, and PerturbEmbedding can all be formulated in Table \ref{tab:methods}. More importantly, we find that PerturbEdge, PerturbNode, and PerturbWeight are actually special cases of PerturbEmbedding, and thus can be expressed in a uniform framework. We provide the equivalent operation of PerturbEmbedding for each strategy below.

\begin{table}[t]
    % \small
    \centering
    \caption{Overview of different perturb strategies in a view of Bernoulli sampling process.}
    \label{tab:methods}
    \begin{tabular}{cc}
    \toprule
    Strategy & Formula \\
    \midrule
    PerturbEdge & $\mathbf{A} = \mathbf{A} + \triangle_{\mathbf{A}}$ \\
    PerturbNode & $\mathbf{X} = \mathbf{X} + \triangle_{\mathbf{X}}$ \\
    PerturbWeight & $\mathbf{W} = \mathbf{W} + \triangle_{\mathbf{W}}$ \\
    PerturbEmbedding & $\mathbf{H} = \mathbf{H} + \triangle_{\mathbf{H}}$ \\
    \bottomrule
    \footnotesize
    % $s.t.\ \epsilon\sim Bernoulli(1-\delta)$
    \end{tabular}
    % \normalsize
\end{table}

\emph{PerturbEdge.} The earliest and most primitive way of perturbing the graph is to randomly drop edges \cite{dai2018adversarial}. The joint training of such cheap adversarial perturbations is shown to slightly improve the robustness of standard GNN models towards both graph and node classiﬁcation tasks. Taking Message-Passing GNNs as an Example, applying perturbations $\triangle_{\mathbf{A}}$ on Edge $\mathbf{A}$ is equivalent to directly applying perturbations $\triangle_{\mathbf{H}}$ on embedding $\mathbf{H}^{(0)}$ at the 1-th layer, which
blends the information of both node features and the graph:
\begin{align}
\label{perturbEdge}
\mathbf{H}^{(0)} = (\mathbf{A}+\triangle_{\mathbf{A}})\mathbf{X}\mathbf{W^{(0)}} = \mathbf{A}\mathbf{X}\mathbf{W^{(0)}} + \mathbf{\triangle_{\mathbf{A}}}\mathbf{X}\mathbf{W^{(0)}} =  \mathbf{H}^{(0)} + \mathbf{\triangle_{\mathbf{H}}}
\end{align}
where $\triangle_{\mathbf{A}} = \{{\triangle_{\mathbf{A}}}_{ij} \in \{0, -1\} \land {\triangle_{\mathbf{A}}}_{ij} \in \mathbf{A}\}$ indicates the edges that need to be dropped.

\emph{PerturbNode.} 
% Besides links, among the series of defensive approaches, the adversarial training based on feature perturbation generally has achieved superior robustness than others \cite{xue2021cap}. The maliciously perturbed features could be treated as the augmentation data, which helps model extrapolate the out-of-distribution testing data and alleviate the overfitting \cite{xue2021cap}.
In addition to edges, among various defensive strategies, adversarial training based on feature perturbation generally exhibits greater robustness compared to other methods \cite{xue2021cap}. The intentionally perturbed features can be considered as augmented data, which aids the model in extrapolating to out-of-distribution testing data and mitigates overfitting \cite{xue2021cap}.
Taking Message-Passing GNNs as an Example, applying perturbations $\triangle_{\mathbf{X}} \in R^{N \times F}$ on node feature $\mathbf{X}$ is equivalent to directly applying perturbations $\triangle_{\mathbf{H}}$ on embedding $\mathbf{H}^{(0)}$ at the $1$-th layer, which
blends the information of both node features and the graph:
\begin{align}
\mathbf{H}^{(0)} = \mathbf{A}(\mathbf{X}+\triangle_{\mathbf{X}})\mathbf{W^{(0)}} = \mathbf{A}\mathbf{X}\mathbf{W^{(0)}} + \mathbf{A}\mathbf{\triangle_{\mathbf{X}}}\mathbf{W^{(0)}} = \mathbf{H}^{(0)} + \mathbf{\triangle_{\mathbf{H}}}
\end{align}

% \emph{PerturbWeight.} Adversarial weight perturbation \cite{wu2020adversarial} (AWP) aims to minimize the worst-
% case loss within the small region centered at the model weights, and it seeks to obtain
% parameter whose neighborhood regions have low training loss.

\emph{PerturbWeight}. Adversarial weight perturbation \cite{wu2020adversarial} (AWP) aims to minimize the worst-case loss within a small region centered at the model weights, seeking parameters whose neighborhood regions exhibit low training loss.
In this way, the weight loss landscape has smaller curvature at the
final learned weights, which in turn shrinks the generalization gap. Applying perturbations $\triangle_{\mathbf{W}}$ on Weight $\mathbf{W}^{(l)}$ is equivalent to directly applying perturbations $\triangle_{\mathbf{H}}$ on embedding $\mathbf{H}^{(l+1)}$ at the $(l+1)$-th layer:
\begin{align}
\mathbf{H}^{(l+1)} = \mathbf{A}\mathbf{H}^{(l)}(\mathbf{W}^{(l)}+\triangle_{\mathbf{W}}) &= \mathbf{A}\mathbf{H}^{(l)}\mathbf{W}^{(l)} + \mathbf{A}\mathbf{H}^{(l)}\mathbf{\triangle_{\mathbf{W}}} \nonumber \\ 
&= \mathbf{H}^{(l+1)} + \mathbf{\triangle_{\mathbf{H}}}
\end{align}

\emph{PerturbEmbedding.} LAT-GCN \cite{jin2019latent} proposes injecting adversarial perturbations only into the first layer $\mathbf{H}^{(0)}$ of GCN, leading to indirect perturbations of the graph, which implicitly enforces robustness against structural attacks. However, unlike LAT-GCN, which only injects adversarial perturbations into the first layer of GCN, PerturbEmbedding applies perturbations $\triangle_{\mathbf{H}}$ on embedding $\mathbf{H}^{(l)}$ at every layer for various types of GNNs:
\begin{align}
\mathbf{H}^{(l)} = \mathbf{H}^{(l)} + \mathbf{\triangle_{\mathbf{H}}}
\end{align}
% PerturbEmbedding can be applied to deeper GNNs.

Based on the above descriptions, we find that PerturbEdge, PerturbNode, and PerturbWeight are equivalent to PerturbEmbedding, and they can be regarded as special cases of PerturbEmbedding, which makes PerturbEmbedding the most flexible strategy.

\subsection{Unified random perturbation (non-targeted) for GNNs}
Motivated by \cite{yu2022graph} which adds directed random noises to the embedding for different data, it can smoothly adjust the uniformity of learned embeddings and has been proven to be a very effective data augmentation method for recommendation. Here we also add random noise directly to the graph to generate random perturbations, then try to minimize the loss by learning the more robust graph model parameters $\theta$ under the random perturbations, which is different from previous adversarial training methods which tries to maximize task-oriented loss by generating adversarial perturbations. In this way, the learned GNNs are expected to be resistant to future adversarial attacks. We consider adding random noise/perturbation to the node features $\mathbf{X}$, edges $\mathbf{A}$, weight $\mathbf{w}$ and embedding $\mathbf{H}$ to generate perturbations.

\textbf{Random node perturbation.} Here we directly add random noises to node feature to generate perturbations during training, then train neural networks by incorporating the random feature perturbations on the inputs as follows:
% \begin{align}
%     \mathbf{H^{(0)}} = f(\mathbf{X} + \triangle_{{\mathbf{X}}}, \mathbf{A}) \quad s.t. \quad \lVert \triangle_{\mathbf{X}} \rVert  \le \delta_{\mathbf{X}}
% \end{align}
\begin{align}
\label{random_feature}
     \min_{\theta} \quad L\left(\mathbf{X} + \triangle_{{\mathbf{X}}}, \mathbf{A}, \mathbf{W},\mathbf{H}\right)  \quad s.t. \quad \lVert \triangle_{\mathbf{X}} \rVert_p  \le \delta_{\mathbf{X}}
\end{align}
where $\theta$ is the parameters of GNNs, $\mathbf{H}$ is the hidden representation, $\triangle_{\mathbf{X}}$ denotes the random perturbation bounded by the $\delta_{\mathbf{X}}$ norm ball. The optimization of Eq. \ref{random_feature} seeks to find the optimal weights having the smaller losses at the neighborhood regions centered at the input features, the maliciously perturbed features could also be treated as the augmentation data, which helps the model extrapolate the out-of-distribution testing data and alleviate the overfitting.

\textbf{Random edge perturbation.} We apply the earliest and most primitive way to perturb the graph structure -- randomly drop edges. To be specific, we inject random perturbations to the original graph while training GNNs: 
% \begin{align}
%     \mathbf{H^{(0)}} = f(\mathbf{X}, \mathbf{A} + \triangle_{{\mathbf{A}}}) \quad s.t. \quad\lVert \triangle_{\mathbf{A}} \rVert  \le \delta_{\mathbf{A}}         \\
%     \mathbf{H^{(l)}} = f(\mathbf{H^{(l-1)}}, \mathbf{A} + \triangle_{{\mathbf{A}}} ) \quad s.t. \quad \lVert \triangle_{\mathbf{A}} \rVert  \le \delta_{\mathbf{A}}
% \end{align}
\begin{align}
\label{random_edge}
     \min_{\theta} \quad L\left(\mathbf{X}, \mathbf{A}  + \triangle_{{\mathbf{A}}}, \mathbf{W}, \mathbf{H}\right)
\end{align}
where $\triangle_{\mathbf{A}} = \{{\triangle_{\mathbf{A}}}_{ij} \in \{0, -1\} \land {\triangle_{\mathbf{A}}}_{ij} \in A\}$, the random perturbations are expected to be small, unnoticeable and should not corrupt the majority of graph structures. The joint training of such cheap perturbations is
shown to slightly improve the robustness of standard GNN models towards graph and node classiﬁcation tasks\cite{sun2022adversarial}.

\textbf{Random weight perturbation.} We directly inject random perturbations to weight $\mathbf{W}$ while training GNNs:
\begin{align}
\label{random_weight}
     \min_{\theta} \quad L\left(\mathbf{X}, \mathbf{A}, \mathbf{W} + \triangle_{{\mathbf{W}}},\mathbf{H}\right)  \quad s.t. \quad \lVert \triangle_{\mathbf{W}} \rVert_p  \le \delta_{\mathbf{W}}
\end{align}
where $\triangle_{\mathbf{W}}$ denotes the random perturbation bounded by the $\delta_{\mathbf{W}}$ norm ball, Eq. \ref{random_weight} seeks to obtain parameter $\mathbf{W}$ whose neighborhood regions have low training loss. Therefore, the weight loss landscape is expected to be smoothed to reduce the generalization gap.

\textbf{Random Embedding perturbation.} We inject random perturbations to hidden embedding $\mathbf{H}$ of GNNs, specifically, we inject random perturbations to every hidden layer while training GNNs and minimize the loss under the random embedding perturbations:
% \begin{align}
%     \mathbf{H^{(l)}} = f(\mathbf{H^{(l-1)}} + \triangle_{{\mathbf{H}}} , \mathbf{A} + \triangle_{{\mathbf{A}}} ) \quad s.t. \quad \lVert \triangle_{\mathbf{H}} \rVert  \le \delta_{\mathbf{H}}        
% \end{align}
\begin{align}
\label{random_representation}
     \min_{\theta} \quad L\left(\mathbf{X}, \mathbf{A}, \mathbf{W}, \mathbf{H}   + \triangle_{{\mathbf{H}}}\right)  \quad s.t. \quad \lVert \triangle_{\mathbf{H}} \rVert_p  \le \delta_{\mathbf{H}}
\end{align}
where $\triangle_{\mathbf{H}}$ denotes the random embedding perturbation bounded by the $\delta_{\mathbf{H}}$ norm ball. Hidden layer blends the information of both node features, the graph and weight, injecting adversarial perturbations to hidden embedding  $\mathbf{H}$ indirectly represent the perturbations in $\mathbf{X}$, $\mathbf{G}$ and $\mathbf{W}$, which is more flexible compared with other methods and implicitly enforces robustness to structural attacks.

% \textbf{Unified random noise/perturbation.} Since above single type of perturbation strategy is only suitable on specific datasets and GNNs, We propose unifying random noise/perturbation in GNNs, which is suitable for most cases. Specifically, we add random noise/perturbation to the features, edges and representation to generate perturbations while training GNNs, mathematically, the min-max objective is:

% \begin{align}
% \label{random_unified}
%     \min_{\theta} \quad L\left(f_{\theta}\left(\mathbf{X} + \triangle_{{\mathbf{X}}}, \mathbf{A} + \triangle_{{\mathbf{A}}},\mathbf{H^} + \triangle_{{\mathbf{H}}}\right), y \right) \\ 
%     s.t. \quad \lVert \triangle_{\mathbf{X}} \rVert_p  \le \delta_{\mathbf{X}}, \quad \lVert \triangle_{\mathbf{A}} \rVert_p  \le \delta_{\mathbf{A}}, \quad \lVert \triangle_{\mathbf{H}} \rVert_p  \le \delta_{\mathbf{H}} \nonumber
% \end{align}
% as the perturbations is randomly generated, the above optimization problem is easy to solve. We 

\subsection{Unified adversarial perturbation (targeted) for GNNs}
\begin{table*}[htbp]
    \centering
    \caption{Comparison results of different perturb strategies. 
    The best results are in bold.
    }
    \label{tab:result_overall}
    \resizebox{1\textwidth}{!}{
    \begin{tabular}{c|c|cccc|ccc}
    \toprule[2.0pt]
    % \multirow{2}{*}{ {Model}{Task \& Dataset}} & \multicolumn{7}{c|||}{Node classification} & \multicolumn{3}{c}{Link prediction}\\
    % \cmidrule{2-8} \cmidrule{9-11}
    % & \multicolumn{4}{c}{Heterophilous datasets} & \multicolumn{3}{c}{Homophilous datasets}\\
    % \cmidrule{2-8}
    & & Penn94 & Chameleon & Squirrel & Film & Cora & CiteSeer & PubMed \\
    \midrule[2.0pt]
   & GCN & $82.10 \std{0.34} $  & $64.56 \std{2.85}$ & $50.72\std{0.69}$ & $29.14\std{0.62}$ & $87.28\std{1.26}$ & $76.68 \std{1.64}$ & $87.38\std{1.66}$ \\
     %\cline{2-9}
   % \cmidrule{1-9}
    \multirow{4}{*}{
    \begin{tabular}[c]{@{}c@{}} Random (non-targeted)
    \end{tabular}
    } 

    & GCN-PerturbNode & $82.80\std{0.22}$ & $66.45\std{2.85}$ & $51.59\std{1.63}$ & $29.78\std{1.16}$ & $88.09\std{0.44}$ & $78.16\std{1.59}$ & $\textbf{88.22}\std{\textbf{0.22}}$ \\
   
    & GCN-PerturbEdge & $82.31\std{0.35}$ & $65.83\std{1.77}$ & $51.76\std{1.77}$ & $29.50\std{1.03}$ & $87.73\std{1.49}$ & $\textbf{78.49}\std{\textbf{1.45}}$ & $87.64\std{0.41}$\\
    & GCN-PerturbWeight &
    $82.65 \std{ 0.40}$ &
    $66.36 \std{ 2.03}$ &
    $\textbf{53.12} \std{ \textbf{2.73}}$ &
    $\textbf{30.66} \std{ \textbf{1.33}}$ &
    $87.57 \std{ 1.13}$ &
    $77.87 \std{ 1.26}$ &
    $87.47 \std{ 0.43}$ \\
    & GCN-PerturbEmbedding & $\textbf{82.98}\std{\textbf{0.21}}$ & $\textbf{67.85}\std{\textbf{2.16}}$ & $51.12\std{1.48}$ & $30.25\std{1.48}$ & $\textbf{88.13}\std{\textbf{0.79}}$ & $78.02\std{1.68}$ & $87.99\std{0.33}$\\
   \cmidrule{1-9}
    \multirow{4}{*}{
    \begin{tabular}[c]{@{}c@{}} Adversarial (Targeted)
    \end{tabular}
    } 
    & GCN-PerturbNode & $82.52 \std{0.28}$ & $68.16 \std{1.97}$ & $53.68 \std{1.63}$ & $29.61 \std{0.85}$ & $87.73 \std{0.38}$ & $78.11 \std{1.39}$ & $\textbf{88.35} \std{\textbf{0.56}}$ \\
    & GCN-PerturbEdge & $82.18 \std{0.44}$ & $66.45 \std{1.97}$ & $53.20 \std{1.93}$ & $29.63 \std{1.20}$ & $87.44 \std{1.10}$ & $78.02 \std{1.42}$ & $87.74 \std{0.39}$\\
    & GCN-PerturbWeight &
    $82.47 \std{ 0.34}$ &
    $67.68 \std{ 1.52}$ &
    $51.91 \std{ 2.64}$ &
    $\textbf{29.87} \std{ \textbf{1.76}}$ &
    $87.61 \std{ 1.17}$ &
    $78.27 \std {1.86} $ &
    $87.77 \std{ 0.23}$ \\
    & GCN-PerturbEmbedding & $\textbf{82.72} \std{\textbf{0.37}}$ & $\textbf{68.51} \std{\textbf{1.15}}$ & $\textbf{56.52} \std{\textbf{1.46}}$ & $29.38 \std{1.91}$ & $\textbf{87.81} \std{\textbf{1.38}}$ & $\textbf{78.63} \std{\textbf{1.27}}$ & $87.90 \std{0.27}$\\

        \midrule[2.0pt]
   & GAT & $81.32 \std{0.76}$  & $68.99 \std{1.15}$ & $61.92 \std{2.61}$ & $28.88 \std{1.67}$ & $87.57 \std{1.19}$ & $75.46 \std{1.72}$ & $87.04 \std{0.43}$ \\
   % \cmidrule{1-9}
    \multirow{4}{*}{
    \begin{tabular}[c]{@{}c@{}} Random (non-targeted)
    \end{tabular}
    } 
    & GAT-PerturbNode & $82.10 \std{0.33}$ & $71.32 \std{1.85}$ & $62.32 \std{0.79}$ & $29.88 \std{1.22}$ & $87.73 \std{0.78}$ & $76.48 \std{0.75}$ & $87.54 \std{0.24}$ \\
    & GAT-PerturbEdge & $81.50 \std{0.57}$ & $71.36 \std{1.81}$ & $62.25 \std{1.59}$ & $29.59 \std{2.00}$ & $87.65 \std{1.12}$ & $76.04 \std{1.93}$ & $87.22 \std{0.44}$\\
    & GAT-PerturbWeight &
    $81.57 \std{ 0.81}$ &
    $71.01 \std{ 1.53}$ &
    $62.69 \std{ 1.62}$ &
    $29.84 \std{ 1.93}$ &
    $87.69 \std{ 1.67}$ &
    $76.15 \std{ 1.37}$ &
    $87.16 \std{ 0.16}$ \\
    & GAT-PerturbEmbedding & $\textbf{82.28} \std{\textbf{0.42}}$ & $\textbf{71.49} \std{\textbf{1.57}}$ & $\textbf{62.86} \std{\textbf{0.68}}$ & $\textbf{29.99} \std{\textbf{0.36}}$ & $\textbf{88.01} \std{\textbf{0.39}}$ & $\textbf{77.05} \std{\textbf{0.64}}$ & $\textbf{87.57} \std{\textbf{0.41}}$\\

   \cmidrule{1-9}
    \multirow{4}{*}{
    \begin{tabular}[c]{@{}c@{}} Adversarial (Targeted)
    \end{tabular}
    } 
    & GAT-PerturbNode & $81.33 \std{0.47}$ & $71.23 \std{1.29}$ & $63.11 \std{1.39}$ & $29.21 \std{1.16}$ & $87.77 \std{1.11}$ & $75.86 \std{1.99}$ & $\textbf{87.38} \std{\textbf{0.72}}$ \\
    & GAT-PerturbEdge & $81.45 \std{0.30}$ & $69.82 \std{1.16}$ & $62.07 \std{0.91}$ & $29.11 \std{0.70}$ & $87.73 \std{1.59}$ & $76.28 \std{1.72}$ & $87.15 \std{0.43}$\\
    & GAT-PerturbWeight &
    $81.83 \std{ 0.64} $&
    $70.92 \std{ 0.81} $&
    $62.11 \std{ 1.28} $&
    $29.08 \std{ 1.30} $&
    $ 87.81 \std{ 1.13}$&
    $ 75.82 \std{ 1.63}$&
    $87.08 \std{ 0.31} $\\
    & GAT-PerturbEmbedding & $\textbf{81.85} \std{\textbf{0.80}}$ & $\textbf{71.54} \std{\textbf{2.49}}$ & $\textbf{63.34} \std{\textbf{1.24}}$ & $\textbf{29.34} \std{\textbf{1.86}}$ & $\textbf{87.93} \std{\textbf{1.21}}$ & $\textbf{76.33} \std{\textbf{2.17}}$ & $87.31 \std{0.32}$\\

 \midrule[2.0pt]
       & LINKX & $84.71 \std{0.52}$  & $68.42 \std{1.38}$ & $61.81 \std{1.80}$ & $36.10 \std{1.55}$ & $85.35 \std{0.97}$ & $74.08 \std{1.52}$ & $87.44 \std{0.76}$ \\
   % \cmidrule{1-9}
    \multirow{4}{*}{
    \begin{tabular}[c]{@{}c@{}} Random (non-targeted)
    \end{tabular}
    } 
    & LINKX-PerturbNode & $84.75 \std{0.24}$ & $70.13 \std{0.48}$ & $63.00 \std{0.48}$ & $37.04 \std{0.55}$ & $85.79 \std{1.30}$ & $\textbf{74.76} \std{\textbf{1.54}}$ & $87.73 \std{0.27}$ \\
    & LINKX-PerturbEdge & $84.65 \std{0.42}$ & $70.04 \std{0.86}$ & $62.21 \std{0.2}0$ & $36.96 \std{0.42}$ & $85.47 \std{1.04}$ & $74.49 \std{0.82}$ & $87.72 \std{0.26}$\\
    & LINKX-PerturbWeight &
    $84.54 \std{ 0.29}$ &
    $70.83 \std{ 0.74}$ &
    $63.07 \std{ 0.79}$ &
    $36.86 \std{ 0.82}$ &
    $85.75 \std{ 1.31}$ &
    $73.83 \std{ 1.21}$ &
     $87.47 \std{0.43}$\\
    & LINKX-PerturbEmbedding & $\textbf{85.42} \std{\textbf{0.47}}$ & $\textbf{71.18} \std{\textbf{0.84}}$ & $\textbf{63.73} \std{\textbf{0.49}}$ & $\textbf{37.47} \std{\textbf{0.82}}$ & $\textbf{86.00} \std{\textbf{1.97}}$ & $74.37 \std{1.58}$ & $\textbf{88.22} \std{\textbf{0.27}}$\\

   \cmidrule{1-9}
    \multirow{4}{*}{
    \begin{tabular}[c]{@{}c@{}}  Adversarial (Targeted)
    \end{tabular}
    } 
    & LINKX-PerturbNode & $84.61 \std{0.30}$ & $70.57 \std{0.61}$ & $62.94 \std{0.63}$ & $37.51 \std{0.72}$ & $85.71 \std{1.14}$ & $74.24 \std{1.66}$ & $87.50 \std{0.48}$ \\
    & LINKX-PerturbEdge & $84.74 \std{0.13}$ & $70.57 \std{0.36}$ & $61.96 \std{0.41}$ & $37.57 \std{0.88}$ & $84.91 \std{1.29}$ & $73.74 \std{1.17}$ & $87.47 \std{0.60}$\\
    & LINKX-PerturbWeight &
    $84.84 \std{0.19} $ &
    $ 70.92 \std{ 0.67}$ &
    $63.17 \std{ 0.35} $ &
    $37.38 \std{ 0.91} $ &
    $85.43 \std{ 1.81} $ &
    $74.61 \std{ 0.50} $ &
    $87.64 \std{ 0.11} $\\
    & LINKX-PerturbEmbedding & $\textbf{84.93} \std{ \textbf{0.20}}$ & $\textbf{71.32} \std{\textbf{0.42}}$ & $\textbf{63.32} \std{\textbf{0.78}}$ & $\textbf{37.75} \std{\textbf{0.53}}$ & $\textbf{86.36} \std{\textbf{2.45}}$ & $\textbf{74.70} \std{\textbf{1.81}}$ & $\textbf{88.42} \std{\textbf{0.38}}$\\

    \bottomrule[2.0pt]
    \end{tabular}
    }
\end{table*}

Previous adversarial GNNs perturb graphs with gradient-based adversarial perturbations during training, which is a time-consuming process. we introduce a novel generator for graph structure data, which tries to maximize task-oriented loss by generating targeted adversarial perturbations, and GNNs (discriminator)  tries to minimize the loss by learning the more robust graph model parameters $\theta$ under the targeted adversarial perturbations. \blue{}Specifically, We apply different forms of targeted perturbations to graph, namely features, edges, weight, and representation.

\textbf{Adversarial node perturbation.} We inject perturbations to the node features $\mathbf{X}$ while training GNNs. The adversarial learning objective following the min-max formulation:
\begin{align}
\label{adversarial_feature}
    \min_{\theta} \max_{\beta} &  L\left( \mathbf{X} + \triangle_{\mathbf{X}}, \mathbf{A}, \mathbf{W}, \mathbf{H} \right) \\
    \text{where} \quad \triangle_{\mathbf{X}} &= Generator_{\beta}(\mathbf{X}) \quad \text{s.t.} \quad \lVert \triangle_{\mathbf{X}} \rVert_p \leq \delta_{\mathbf{X}} \nonumber
    % \triangle_{\mathbf{X}} = arg \max_{\triangle_{\mathbf{X}}} L\left(f_{\theta}\left(\mathbf{X} + \triangle_{\mathbf{X}}\right), y \right)
\end{align}

where the generator, with parameters $\beta$, aims to generate worst-case adversarial perturbations $\triangle_{\mathbf{X}}$ to maximize task-oriented loss, while the GNNs minimize the loss under these perturbations.
%where the targeted adversarial perturbations $\triangle_{\mathbf{X}}$ are generated based node feature $\mathbf{X}$, $\beta$ is the model parameters of Generator, the goal of Generator is to cause maximize task-oriented loss, thus generating worst-case adversarial perturbations. GNNs (discriminator) tries to minimize the loss by learning the more robust graph model parameters $\theta$ under the targeted adversarial perturbations. The parameters of the generator and GNNs (discriminator) are updated alternately.

\textbf{Adversarial edge perturbation.} GNNs are sensitive to small perturbations of graph structure \cite{dai2018adversarial}, to defends against perturbations on graph topology, we inject perturbations to the origin graph while training GNNs, The adversarial learning objective following the min-max formulation:
\begin{align}
\label{adversarial_edge}
     \min_{\theta} \max_{\beta} &L\left(\mathbf{X}, \mathbf{A} + \triangle_{\mathbf{A}}, \mathbf{W}, \mathbf{H}\right) \\
    where \quad \triangle_{\mathbf{A}} = Generat&or_{\beta}(\mathbf{A})  \quad s.t. \quad \lVert \triangle_{\mathbf{A}} \rVert_p  \le \delta_{\mathbf{A}} \nonumber
      % \triangle_{\mathbf{X}} = arg \max_{\triangle_{\mathbf{X}}} L\left(f_{\theta}\left(\mathbf{X} + \triangle_{\mathbf{X}}\right), y \right)
\end{align}
where the generator creates targeted adversarial perturbations $\triangle_{\mathbf{A}}$ based on the adjacency matrix $\mathbf{A}$, using a learnable MLP to generate a soft mask matrix $\mathbf{M}$, which is then used to obtain $\triangle_{\mathbf{A}}$.
%where $\triangle_{\mathbf{A}} = \{{\triangle_{\mathbf{A}}}_{ij} \in \{0, -1\} \land {\triangle_{\mathbf{A}}}_{ij} \in A\}$, the targeted adversarial perturbations $\triangle_{\mathbf{A}}$ are generated based node feature $\mathbf{A}$, $\beta$ is the model parameters of Generator. However, directly generating adversarial perturbations $\triangle_{\mathbf{A}}$ is challenging due to discrete nature of the graph perturbations. A common strategy is to use a binary mask matrix $\mathbf{M} = \{0, 1\}^{n \times n}$ on the adjacency matrix $A$. We adopt a learnable MLP to generate a soft mask matrix $\mathbf{M} = \{0, 1\}^{n \times n}$ as follows:
\begin{align}
    \mathbf{M}_{i,j} = {\mathbf{Z}_{i}^{(m)}}^{T} \cdot {\mathbf{Z}_{j}^{(m)}}, \mathbf{Z}^{(m)} = \mathbf{MLP}_{\beta}(A)
\end{align}
where $\mathbf{Z}^{(m)}$ is the node representation. Then, we obtain the targeted adversarial perturbations $\triangle_{\mathbf{A}}$:
\begin{align}
    \triangle_{\mathbf{A}} = -\mathbf{Top}_{t}(\mathbf{M} \odot \mathbf{A})
\end{align}
where $\odot$ means the element-wise matrix multiplication, and $\mathbf{Top}_{t}$ selects the top $t$-percentage of elements with the largest values. In this way, we can efficiently generate adversarial perturbations $\triangle_{\mathbf{A}}$.

\textbf{Adversarial Weight perturbation.} {The adversarial learning objective of following min-max formulation:}

\begin{align}
    \min_{\theta} \max_{\beta} &  L\left( \mathbf{X}, \mathbf{A}, \mathbf{W} + \Delta_{\mathbf{W}}, \mathbf{H} \right) \\
    \text{where} \quad \Delta_{\mathbf{W}} &= \text{Generator}_{\beta}(\mathbf{W}) \quad \text{s.t.} \quad \lVert \Delta_{\mathbf{W}} \rVert_p \leq \delta_{\mathbf{W}} \nonumber
    % \triangle_{\mathbf{X}} = arg \max_{\triangle_{\mathbf{X}}} L\left(f_{\theta}\left(\mathbf{X} + \triangle_{\mathbf{X}}\right), y \right)
\end{align}
where the targeted adversarial perturbations $\triangle_{\mathbf{W}}$ are generated based node embedding $\mathbf{W}$, $\beta$ is the model parameters of Generator.

\textbf{Adversarial Embedding perturbation.} For GNNs, the features of nodes are aggregated by the neighborhoods. Perturbation on the hidden representation output to affect the nodes with their neighborhoods would produce a more challenging view. Specifically, we inject random perturbations to every hidden embedding while training GNNs, we optimize the following min-max formulation:
\begin{align}
\label{adversarial_representation}
    \min_{\theta} \max_{\beta} &  L\left( \mathbf{X}, \mathbf{A}, \mathbf{W}, \mathbf{H} + \triangle_{\mathbf{H}} \right) \\
    \text{where} \quad \triangle_{\mathbf{H}} &= Generator_{\beta}(\mathbf{H}) \quad \text{s.t.} \quad \lVert \triangle_{\mathbf{H}} \rVert_p \leq \delta_{\mathbf{H}} \nonumber
    % \triangle_{\mathbf{X}} = arg \max_{\triangle_{\mathbf{X}}} L\left(f_{\theta}\left(\mathbf{X} + \triangle_{\mathbf{X}}\right), y \right)
\end{align}
where the targeted adversarial perturbations $\triangle_{\mathbf{H}}$ are generated based node embedding $\mathbf{H}$, $\beta$ is the model parameters of Generator. 
%The whole proposed adversarial perturbation algorithm is detailed in Algorithm \ref{alg:pseudo_code}.

The unified adversarial perturbation (targeted) algorithm is detailed in Appendix \ref{alg:Algorithm_details}, which alternates between updating the generator and GNN parameters using stochastic gradient descent.

% \textbf{Unified adversarial noise/perturbation}. We argue that above single type of perturbation strategy is only suitable on specific datasets, specific scenarios and specific GNNs, We propose unifying random noise/perturbation in GNNs, which is suitable for most scenarios. Specifically, we inject targeted noise/perturbation to the features, edges and representation while training GNNs, mathematically, the min-max objective is:
% \begin{align}
% \label{adversarial_unified}
%      \min_{\theta} \max_{\beta} L\left(f_{\theta}\left(\mathbf{X} + \triangle_{\mathbf{X}}, \mathbf{A} - \triangle_{\mathbf{A}}, \mathbf{H} + \triangle_{\mathbf{H}}\right), y \right)
% \end{align}
% \begin{align}
% \label{adversarial_unified2}
%     \triangle_{\mathbf{X}} = Generator_{\beta}(\mathbf{X}), \triangle_{\mathbf{A}} = Generator_{\beta}(\mathbf{A}),\triangle_{\mathbf{H}} = Generator_{\beta}(\mathbf{H}) \\ 
%     \quad s.t. \quad \lVert \triangle_{\mathbf{X}} \rVert_p  \le \delta_{\mathbf{X}}, \lVert \triangle_{\mathbf{H}} \rVert_p  \le \delta_{\mathbf{H}} \nonumber
% \end{align}

% where $\triangle_{\mathbf{A}}$, $\triangle_{\mathbf{X}}$ and $\triangle_{\mathbf{H}}$ are generated by three different generators, and we denote its parameters as $\theta$. During the training process, GNN and the Generator adversarially confront each other, forcing GNN to learn the robust parameters $\theta$ to defend against different forms of perturbations. The whole proposed algorithm of is detailed in Algorithm \ref{alg:pseudo_code}.

\section{Experiments}

\begin{table*}[]
\captionsetup{font={small,stretch=1.25}, labelfont={bf}}
 \renewcommand{\arraystretch}{1}
    \caption{Comparison with other adversarial training GNNs. The best results per dataset is highlighted. (M) denotes out of memory.}
    \label{tab:adversarial_results}
    \centering
        \resizebox{0.99\linewidth}{!}{
    \begin{tabular}{cccccccc}
        \toprule[1.5pt]
&Penn94& Chameleon & Squirrel  & Film  & Cora & Citeseer & Pubmed \\
        
   \toprule[1.0pt]
GCN &$82.10\pm{0.34}$ &$64.56 \std{ 2.85}$ &$50.72 \std{ 0.69}$ & $29.14 \std{ 0.62}$  & $87.28 \std{ 1.26}$ &$76.68 \std{ 1.64}$  & $87.38 \std{ 0.66}$ \\    
GCN + VAT & $55.98 \pm{1.44}$ &$59.17\std{ 3.78}$ &$27.91 \std{ 0.89}$ & $29.14 \std{ 0.62}$  & $\textbf{88.37} \std{ \textbf{1.60}}$ &$78.22 \std{ 1.69}$  & $86.83 \std{ 0.98}$ \\
GCN + ADV\_Train & $72.38 \pm{0.81}$&$59.69 \std{ 1.66}$ &$39.02 \std{ 1.54}$ &$\textbf{30.65} \std{ \textbf{1.16}}$ & $87.81 \std{ 1.67}$  & $78.01 \std{ 1.85}$ &$87.70 \std{ 0.53}$ \\
GCN + TW-AWP & (M) &$47.15 \std{ 6.53}$ &$31.58 \std{ 1.35}$ & $29.45 \std{ 1.33}$  & $87.97 \std{ 0.91}$ &$76.07 \std{ 0.95}$  & $86.28 \std{ 0.98}$ \\
GCN + Flag & (M) &$66.49 \std{ 2.48}$ &$50.10 \std{ 1.02}$ & $29.23 \std{ 0.84}$  & $87.48 \std{ 0.95}$ &$75.48 \std{ 1.31}$  & $87.88 \std{ 0.53}$ \\
\textbf{GCN} \textbf{+} \textbf{PerturbEmbedding} &$\textbf{82.98} \std{\textbf{0.21}} $ &$\textbf{68.51} \std{ \textbf{1.15}}$ &$\textbf{56.52} \std{ \textbf{1.46}}$ & $30.25 \std{ 1.48}$  & $88.13 \std{ 0.79}$ &$\textbf{78.63} \std{ \textbf{1.27}}$  & $\textbf{87.99} \std{ \textbf{0.33}}$ \\
        \toprule[1.5pt]    
    \end{tabular}}
\end{table*}

In this section, we conduct extensive experiments on various datasets to empirically demonstrate {our method}'s effectiveness by answering the following questions:
\begin{itemize}
\item \textbf{(RQ 1)} How does PerturbEmbedding enhance the GNNs in node classification tasks across various datasets and backbone models?
\item \textbf{(RQ 2)} How does PerturbEmbedding perform under different types of perturbations, including random and adversarial, in maintaining the robustness of GNNs?
\item \textbf{(RQ 3)} What are the individual contributions of the components within PerturbEmbedding to its overall performance, and how do they compare to specialized perturbation methods?
\item \textbf{(RQ 4)} How efficient is PerturbEmbedding,  and what are the implications for practical applications?
    % \item \textbf{(RQ 1)} How effective is the proposed \blue{method} applied to current popular GNNs in the node classification tasks?
    % %neighborhood pattern bias?
    % \item \textbf{(RQ 2)}  How effective our proposed method compared with previous method? 
    % \item \textbf{(RQ 3)} How robust our proposed \blue{method} is in the unknown perturbations?
    % \item \textbf{(RQ 4)} How does each component contribute to the performance of our proposed \blue{method}?
\end{itemize}

\subsection{Experimental Setup.}

\textbf{Datasets.} We employ 7 graph datasets in our experiments, including 3 widely used homophilous datasets \emph{Cora}, \emph{Citeseer} and \emph{PubMed} \cite{yang2016revisiting}, and 4 heterophilous datasets \emph{Penn94}, \emph{Chameleon}, \emph{Squirrel} and \emph{Chameleon}. Dataset statistics are in Appendix Table \ref{tab:datasets statistics}.

% namely GCN \cite{kipf2016semi}, GAT\cite{velickovic2017graph} and LINKX\cite{lim2021large}, and compare their performances with PerturbEdge, PerturbNode and PerturbWeight on different datasets.

\textbf{Training and Evaluation.} We conduct experiments on \emph{Cora}, \emph{Citeseer} and \emph{PubMed} with 60/20/20 train/val/test splits, for Penn94, we follow the setting in \cite{lim2021large} with 50/25/25 train/val/test splits, for datasets of Film, Squirrel, Chameleon, we follow the widely used semi-supervised setting in \cite{pei2020geom} with the standard 48/32/20 train/val/test splits. We use classification accuracy as the evaluation metric. For a fair comparison, we conduct five independent runs and report the mean result with the standard deviation over it.

\textbf{Baseline methods.} We compare proposed unified framework PerturbEmbedding with other perturbation strategies, including PerturbEdge, PerturbNode, and PerturbWeight. According to the generation form of perturbations, they can be further divided into random perturbations (non-targeted) and adversarial perturbations (targeted). We combine different perturbation strategies and forms of perturbations, and adopt them on various GNNs as the backbone model, and compare their performances on different datasets.

\textbf{Backbone models.} To illustrate that our method can be applied in different different scenarios and graph neural networks with different structures. we mainly consider three mainstream GNNs as our backbone models: GCN \cite{kipf2016semi}, GAT\cite{velickovic2017graph} and LINKX\cite{lim2021large}. GCN and GAT are the most representative homophilous methods and adopt the message-passing framework. Different from message-passing GNNs that aggregate messages from neighbor nodes for the target node, LINKX\cite{lim2021large} separately embeds node features and adjacency information with MLPs, combines the embeddings by concatenation, which achieves superior performance on heterophilous graphs. For GCN and GAT, we apply PerturbEmbedding to the first layer output $H^{(1)}$ after graph convolution. For LINKX, we apply PerturbEmbedding to every hidden layer.
\subsection{Overall results}
\begin{figure*}[t]
\centering
\begin{subfigure}{0.33\linewidth}
     \includegraphics[scale=0.31]{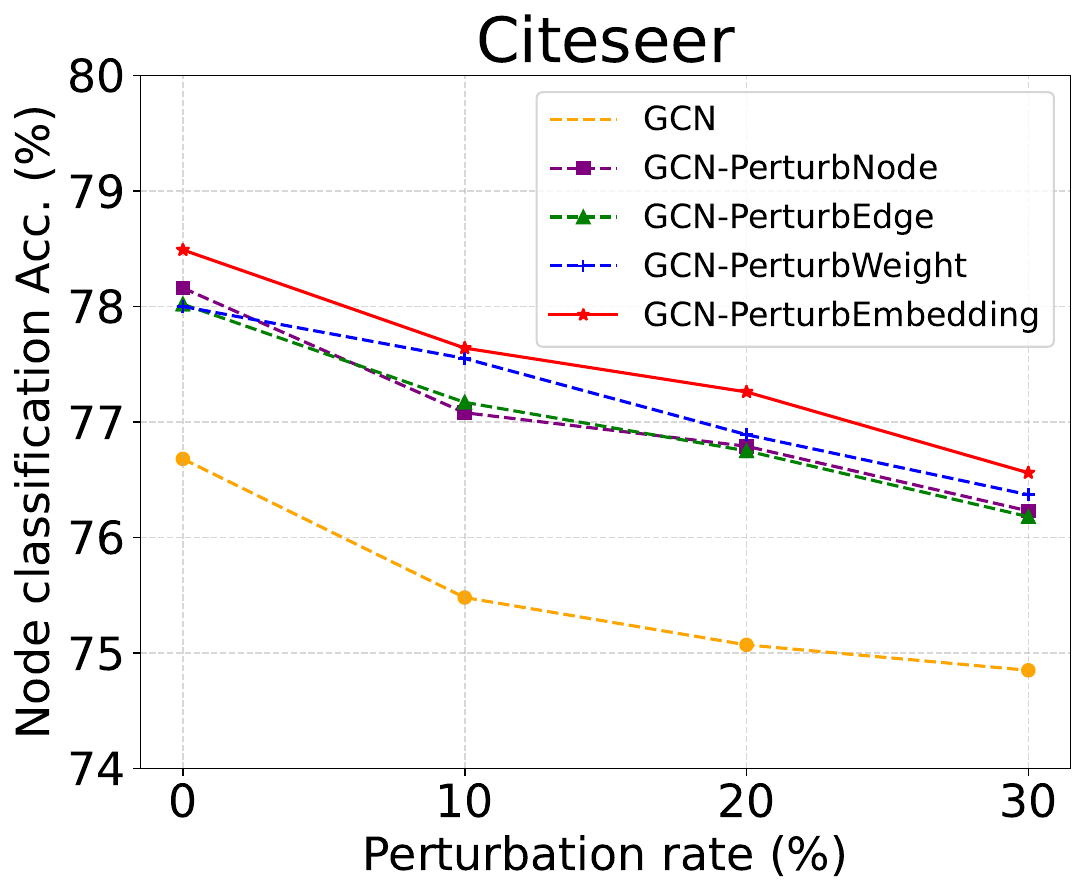}  % 图片路径和大小
\end{subfigure}
\begin{subfigure}{0.33\linewidth}
     \includegraphics[scale=0.31]{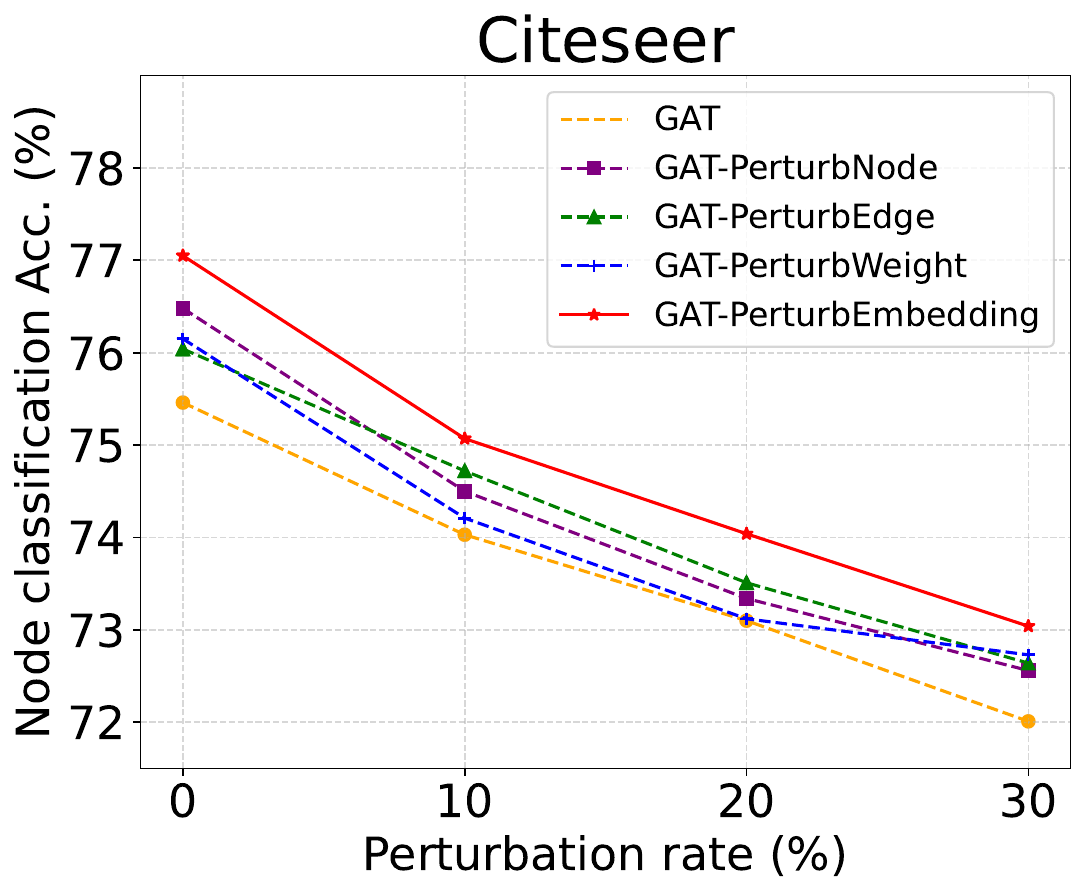}  % 图片路径和大小
\end{subfigure}
\begin{subfigure}{0.33\linewidth}
     \includegraphics[scale=0.31]{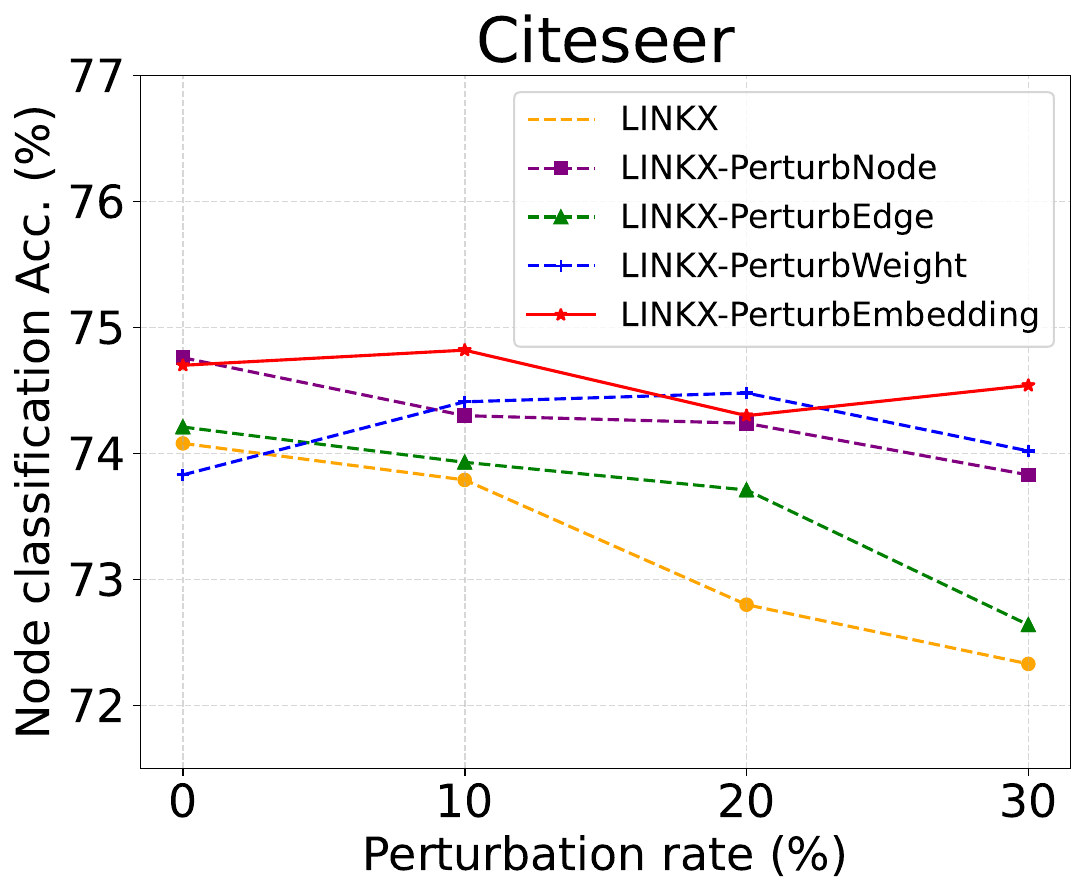}  % 图片路径和大小
\end{subfigure}
\caption{Model performance against graph perturbations. We randomly add a certain ratio of edges into Citeseer.
% and perform the node classification, using GCN, GAT, and LINKX as the backbone. We show the optimal values of random perturbations and adversarial perturbations. 
}    % 整个大图的标题
\label{perturbation}       % 标签
\end{figure*}

\begin{figure*}[t]
\centering
\begin{subfigure}{0.33\linewidth}
     \includegraphics[scale=0.31]{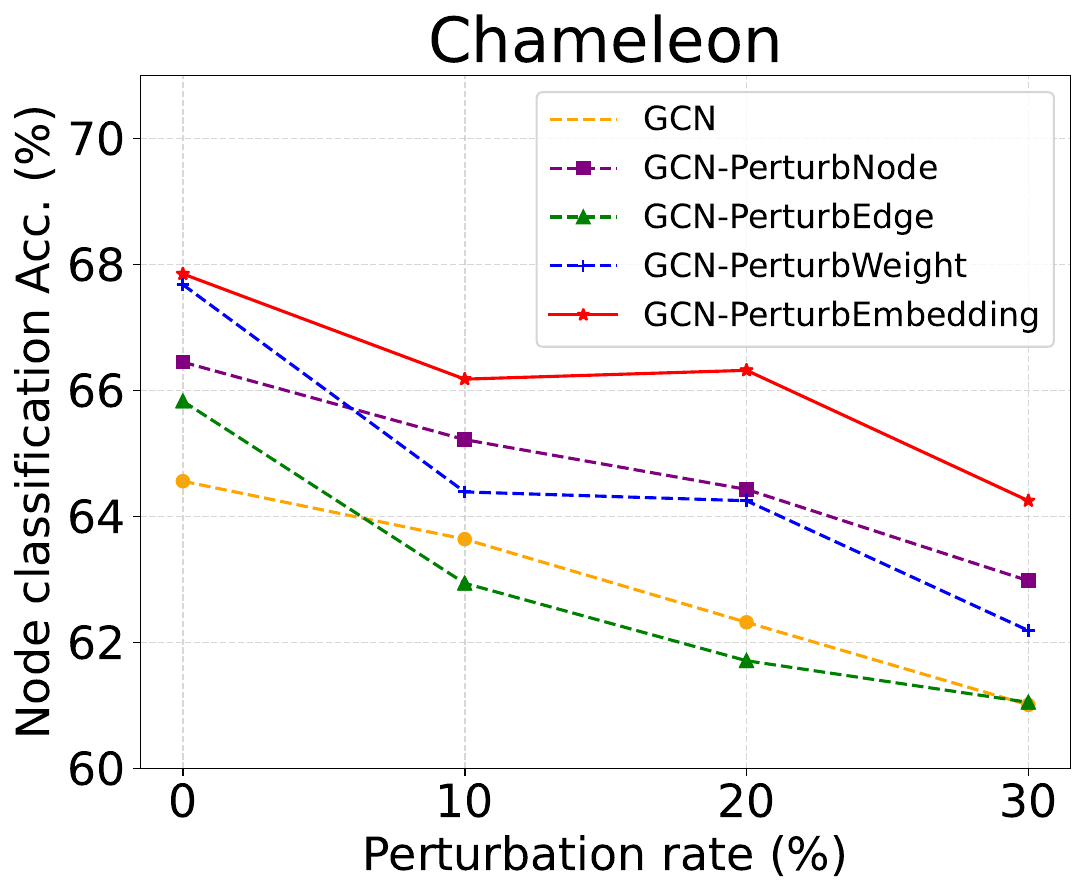}  % 图片路径和大小
\end{subfigure}
\begin{subfigure}{0.33\linewidth}
     \includegraphics[scale=0.31]{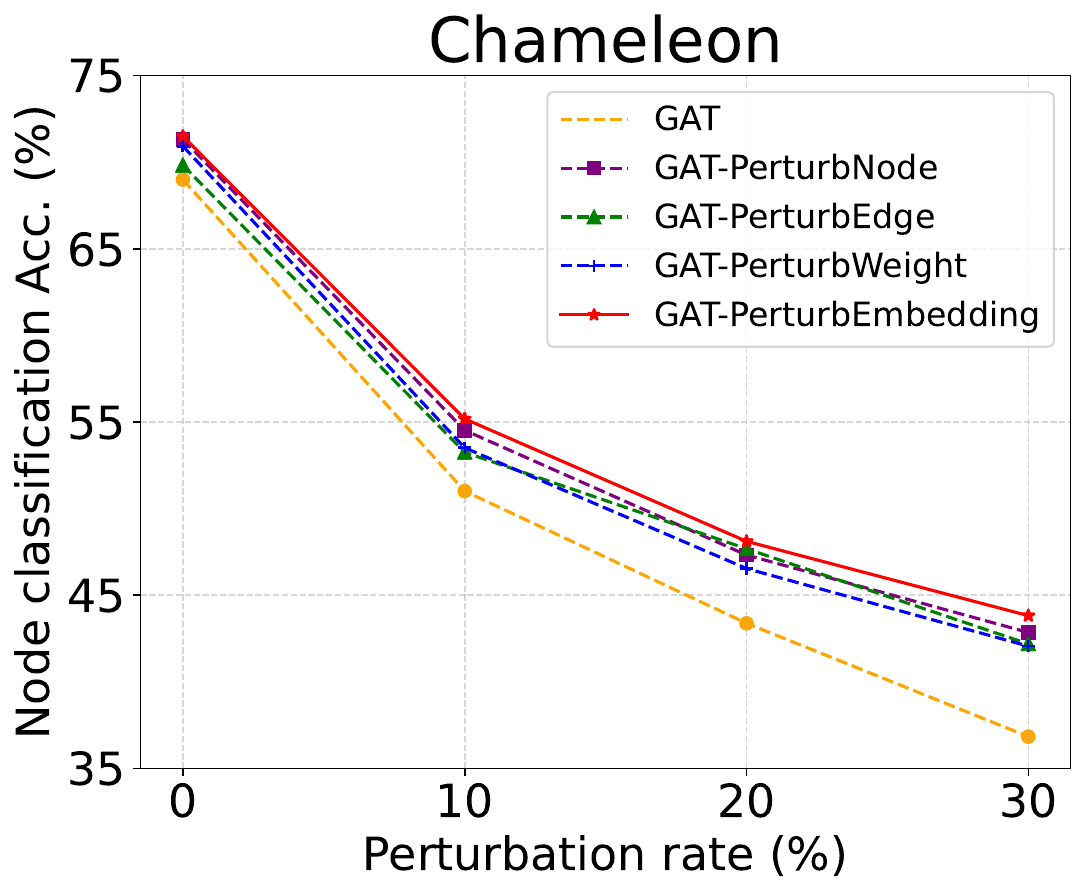}  % 图片路径和大小
\end{subfigure}
\begin{subfigure}{0.33\linewidth}
     \includegraphics[scale=0.31]{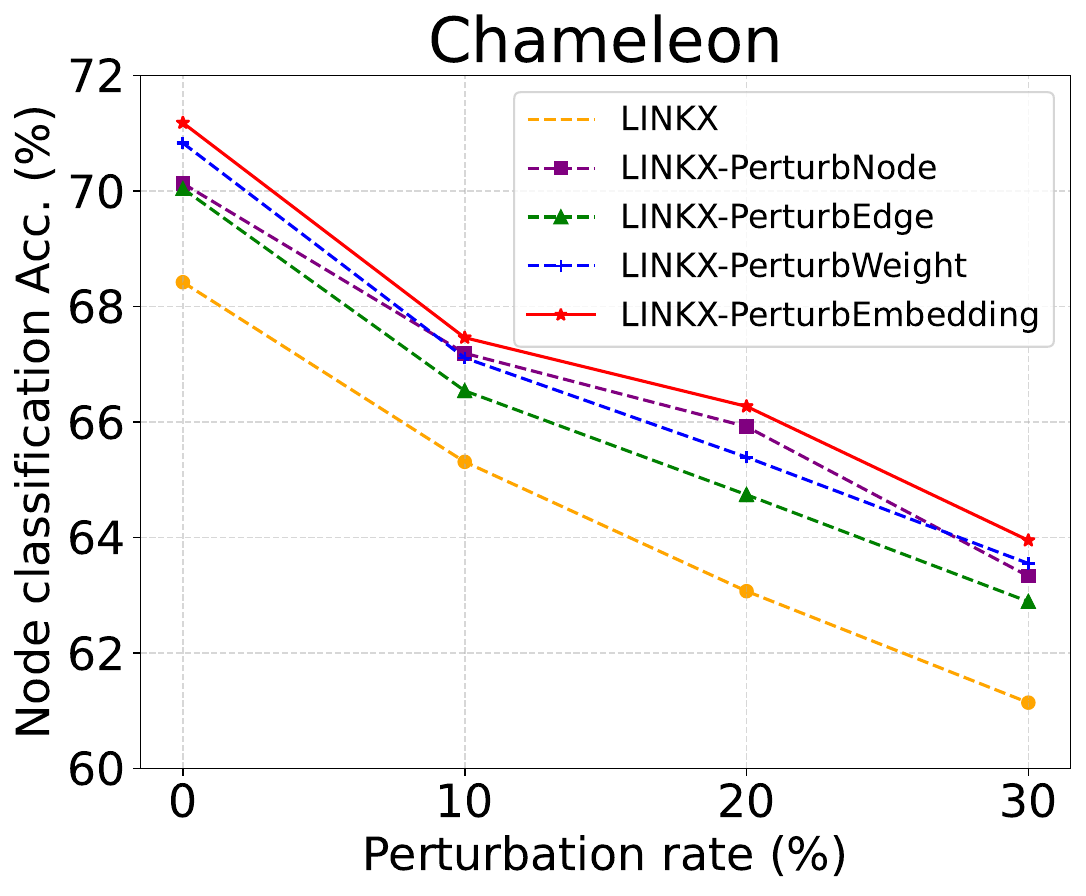}  % 图片路径和大小
\end{subfigure}
\caption{Model performance against graph perturbations. We randomly add a certain ratio of edges into Chameleon.
% and perform the node classification, using GCN, GAT and LINKX as the backbone. We show the optimal values of random perturbations and adversarial perturbations. 
}    % 整个大图的标题
\label{perturbation2}       % 标签
\end{figure*}

\begin{figure*}[t]
\centering
\begin{subfigure}{0.31\linewidth}
     \includegraphics[scale=0.22]{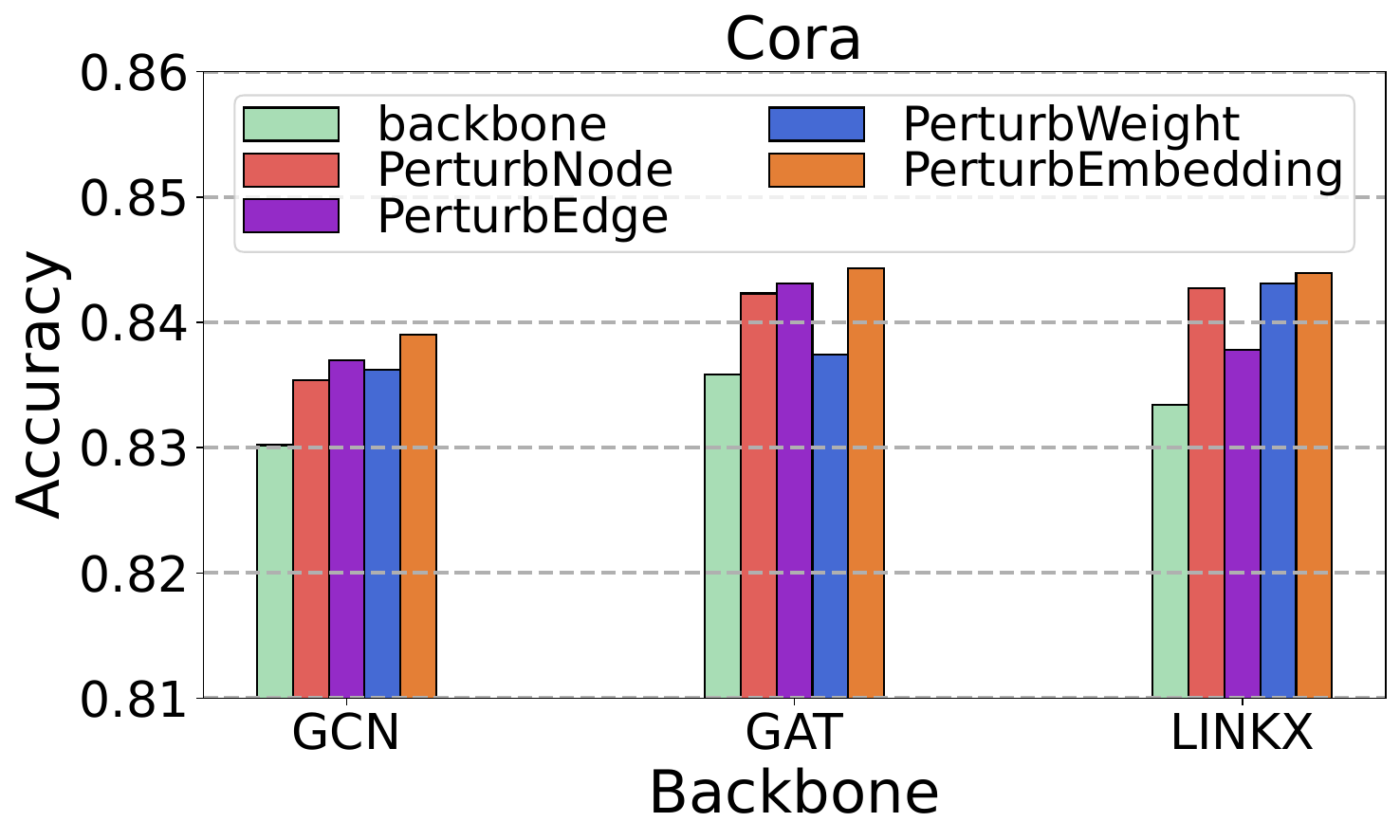}  % 图片路径和大小
\end{subfigure}
\begin{subfigure}{0.31\linewidth}
     \includegraphics[scale=0.22]{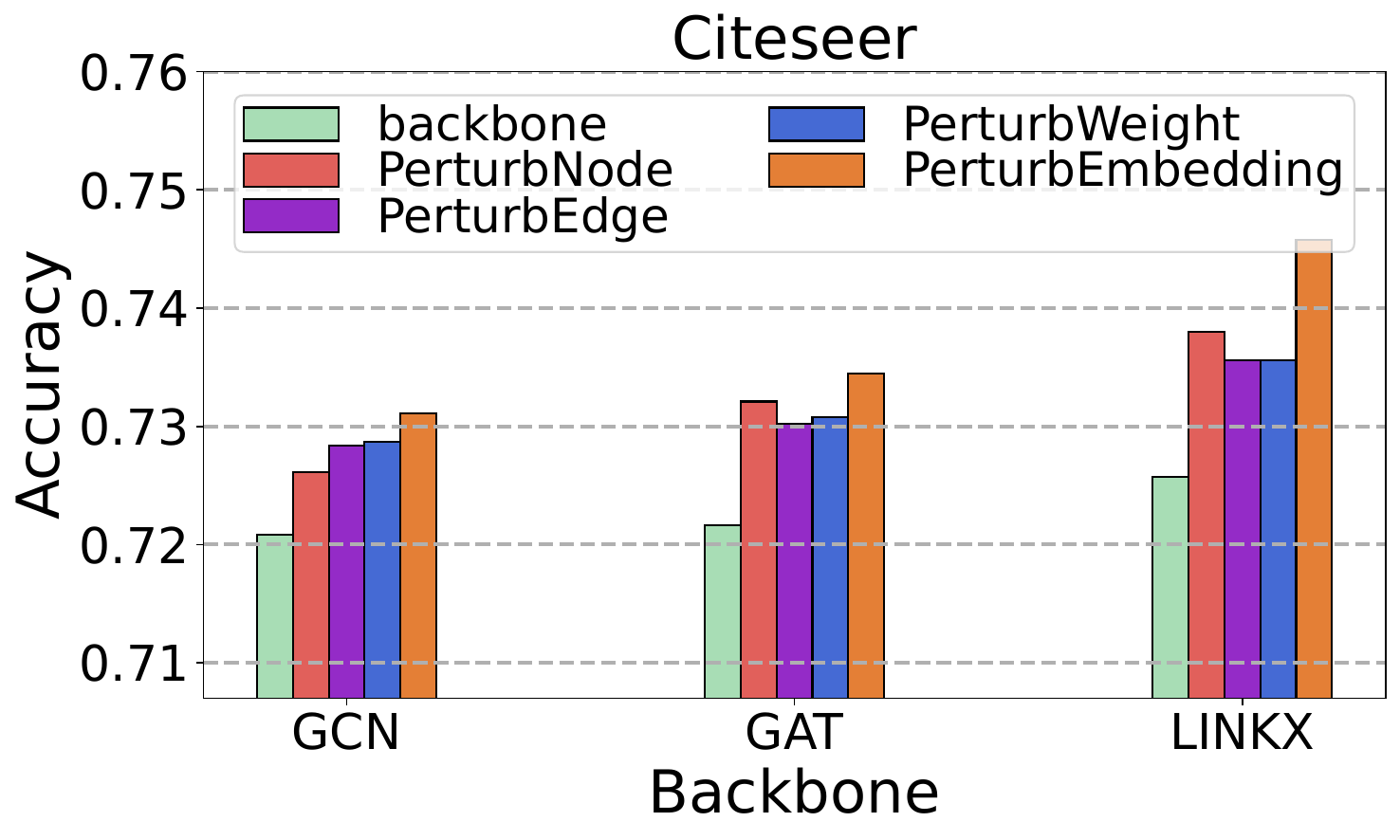}  % 图片路径和大小
\end{subfigure}
\begin{subfigure}{0.31\linewidth}
     \includegraphics[scale=0.22]{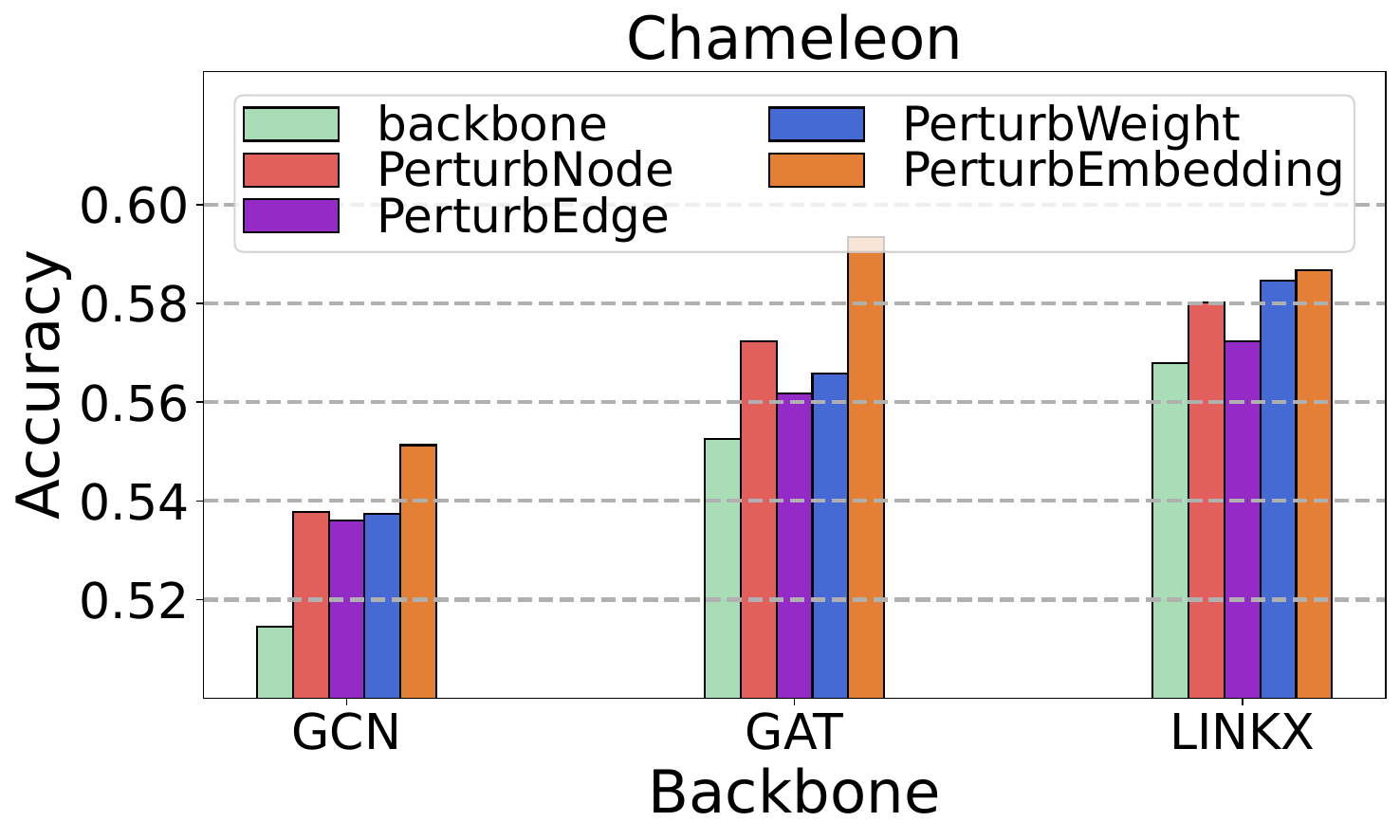}  % 图片路径和大小
\end{subfigure}
\caption{Model Performances Against Meta Attacks (Perturbation rate = 5\%).}    % 整个大图的标题
\label{metaattack}       % 标签
\end{figure*}

% \begin{figure}[]
% \centering
% \begin{subfigure}{0.7\linewidth}
%      \includegraphics[scale=0.23]{photo/Emd_noise_comp_3_98.pdf}  % 图片路径和大小
% \end{subfigure}
% % \begin{subfigure}{0.25\linewidth}
% %      \includegraphics[scale=0.35]{photo/chameleon_Embeddinig_Dist_double_3_98.pdf}  % 图片路径和大小
% % \end{subfigure}
% \caption{Distribution of node representations learned from the dataset of Chameleon. }    % 整个大图的标题
% \label{visualize_chameleon}       % 标签
% \end{figure}

\textbf{To answer RQ 1 adn RQ 2}, we apply our method in different methods, and Table \ref{tab:result_overall} summarizes the overall results.

\textbf{Effect of random/adversarial perturbations.} It is observed that random/adversarial perturbation methods consistently outperform GNNs without random/adversarial perturbations in node classiﬁcation tasks for all 7 datasets. Besides, we see that the effects of random/adversarial perturbation methods vary over different datasets and backbone models.
\begin{itemize}
    \item For example, random perturbation methods on LINKX obtain an average accuracy improvement of {2.1\%} on Chameleon, while {0.35\%} on Pubmed. Meanwhile, random perturbation methods achieve {1.46\%} accuracy improvement for GCN on Citeseer, while only {0.97\%} for GAT.
    \item Adversarial perturbation methods on GCN obtain an average accuracy improvement of {3.1\%} on the chameleon, while {0.48\%} on Film. Meanwhile, adversarial perturbation methods achieve {3.1\%} accuracy improvement for GCN on squirrel, while only {0.85\%} for GAT.
\end{itemize}

\textbf{Comparison of random perturbations and adversarial perturbations.} 
We find that rejecting random perturbations and adversarial perturbations could both improve the performance of the backbone model, and 
the performance of random perturbations and adversarial perturbations is comparable. Taking PerturbEmbedding as the example, we have 21 settings under the node classiﬁcation task, each of which is a combination of different backbone models and datasets. (e.g., GCN-Cora), it is shown that PerturbEmbedding with random perturbations achieves the optimal results in 11 settings while adversarial perturbed get 10 optimal results. which indicates that simply injecting random noise into the embedding during the training process can achieve comparable results.

\textbf{Comparison with other adversarial GNNs}. We compare our method with the following representative adversarial GNNs: 1) \textbf{Attribute Perturbations methods}: Graph Adversarial Training \cite{feng2019graph} (GraphAT), Free Large-scale Adversarial Augmentation on Graphs \cite{kong2020flag} (Flag). 2) \textbf{Structure Perturbations methods}: Adversarial Training \cite{xu2019topology} (ADV\_Train).  3) \textbf{Weight Perturbations methods}: Weighted Truncated Adversarial Weight Perturbation \cite{wu2020adversarial} (WT-AWP) . Table \ref{tab:adversarial_results} shows GCN’s performance with different adversarial augmentations and PerturbEmbedding outperforms all other methods in most datasets, which demonstrates its effectiveness.

\subsection{Robustness analysis} 
\textbf{To answer RQ 3}, We study the robustness of perturbation methods by measuring their ability to handle perturbed graphs. Specifically, we conduct experiments on two settings: 1) \textbf{Random perturbations}: we randomly add a certain ratio of edges into the graph and perform the node classiﬁcation. 2) \textbf{Adversarial attacks}: We also conduct experiments to evaluate the effectiveness of our method against adversarial attacks.

\begin{figure*}[t]
\centering
\begin{subfigure}{0.31\linewidth}
     \includegraphics[scale=0.22]{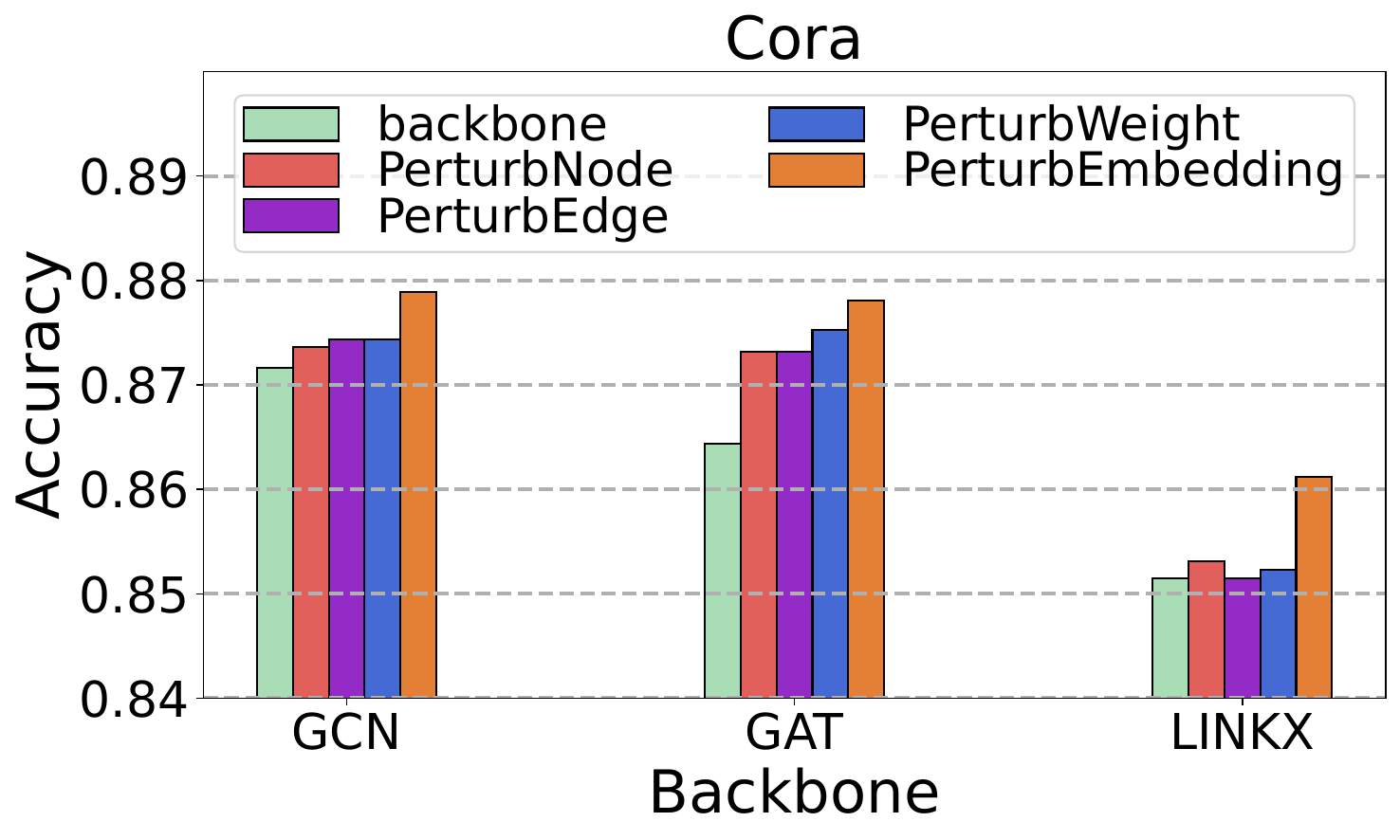}  % 图片路径和大小
\end{subfigure}
\begin{subfigure}{0.31\linewidth}
     \includegraphics[scale=0.22]{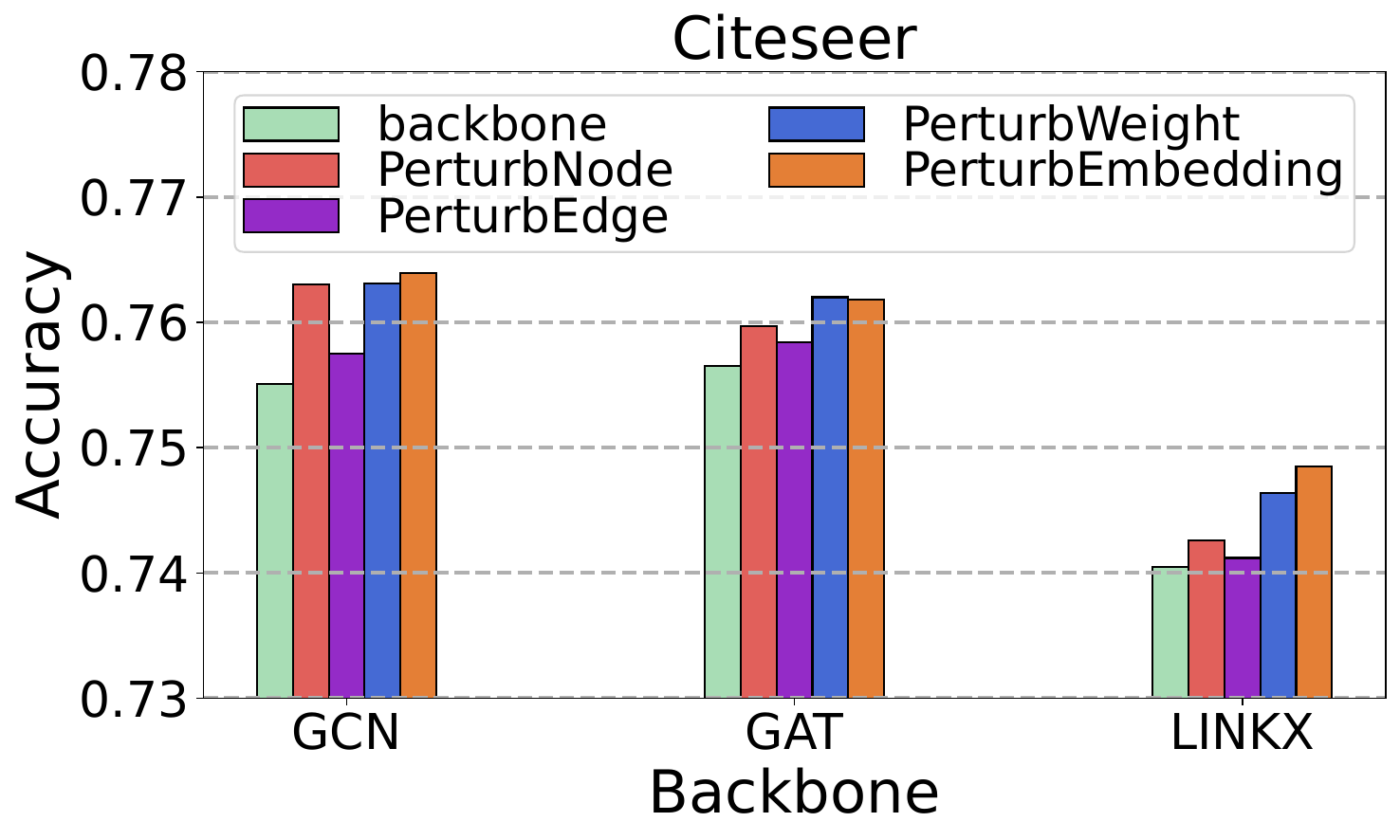}  % 图片路径和大小
\end{subfigure}
\begin{subfigure}{0.31\linewidth}
     \includegraphics[scale=0.22]{photo/chameleon_meta_attack.pdf}  % 图片路径和大小
\end{subfigure}
\caption{Model Performances Against PGD Attacks (Perturbation rate = 5\%).}    % 整个大图的标题
\label{pgdattack}       % 标签
\end{figure*}

\textbf{Random perturbations.} We conduct experiments on citeseer and Chameleon, using GCN and GAT as the backbone model. We randomly add a certain ratio of edges into these datasets and perform the node classiﬁcation, Figure \ref{perturbation} and \ref{perturbation2} show the results. We ﬁnd that all the random perturbation methods have positive effects when the perturbation rate increases from 0\% to \%30, which indicates that the random perturbation methods strengthen the robustness of GNN models. Besides, our proposed PerturbEmbedding shows its versatility and outperforms other perturbation methods in noisy situations.

\begin{figure}[]
\centering 
\includegraphics[width=0.9\linewidth]{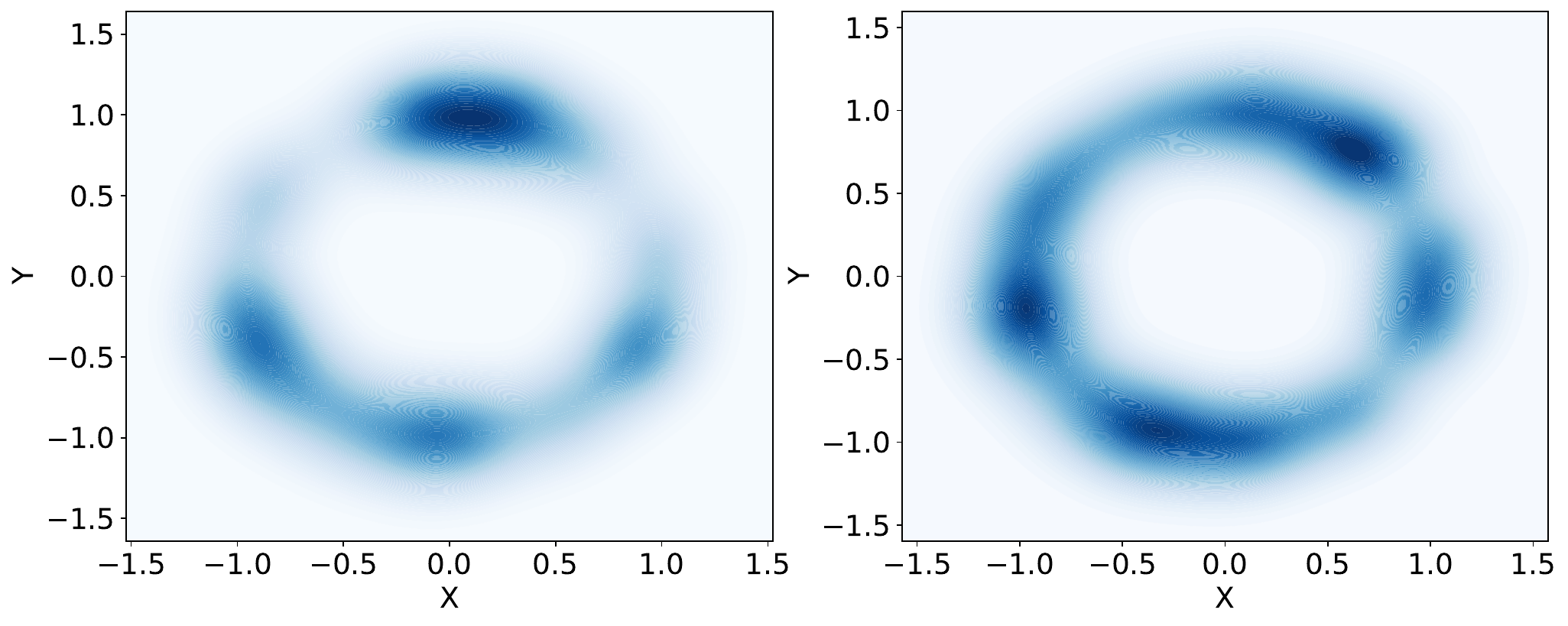}  % 图片路径和大小
\caption{Distribution of node embedding distribution of GCN (left) and our method (right) learned from cora.}    % 整个大图的标题
\label{cora_emb}       % 标签
\end{figure}

\begin{figure}[]
\centering 
\includegraphics[width=0.9\linewidth]{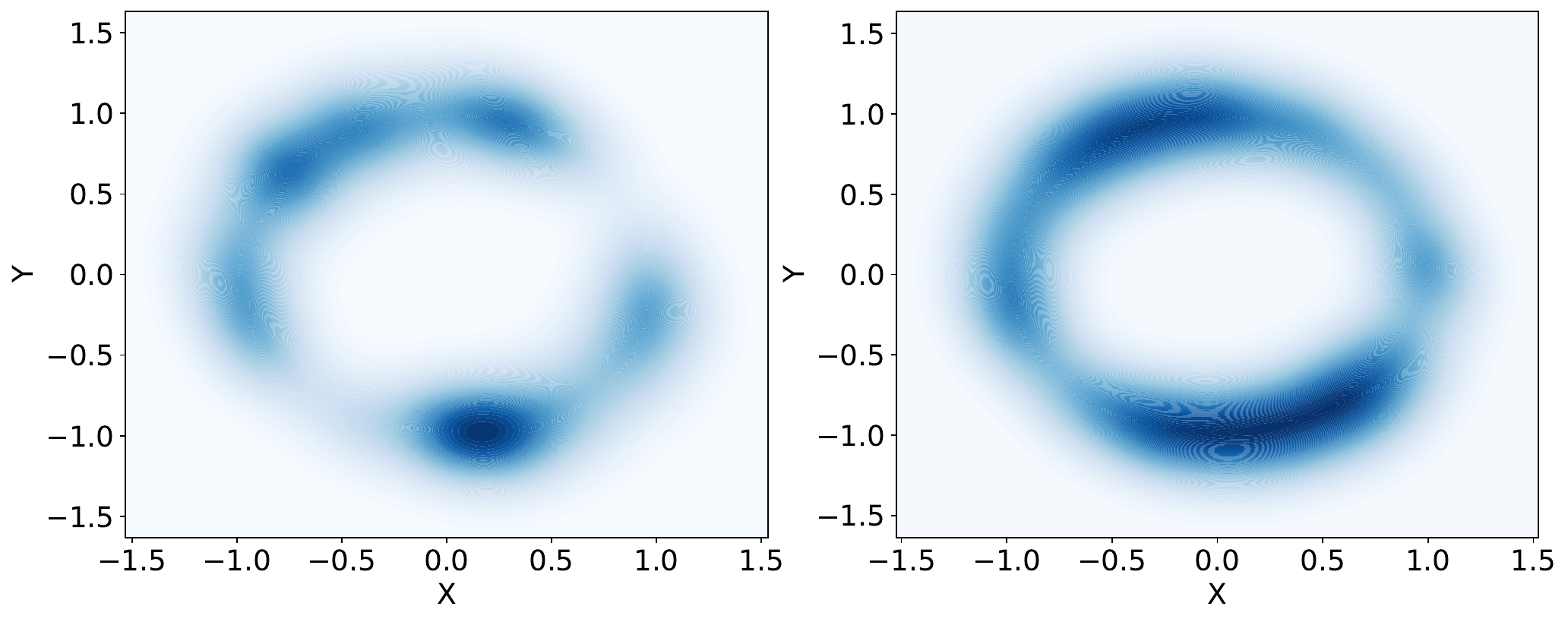}  % 图片路径和大小
\caption{Distribution of node embedding distribution of GCN (left) and our method (right) learned from chameleon.}    % 整个大图的标题
\label{cchameleon_emb}       % 标签
\end{figure}

\textbf{Adversarial Attacks.} We also conduct experiments to evaluate the effectiveness of perturbation methods against adversarial attacks. We apply Meta attacks and PGD attacks to perturb the graph structures on Cora, CiteSeer and Chameleon, using GCN, GAT and LINKX as the backbone model. Figure \ref{metaattack} and \ref{pgdattack} presents the results.

\subsection{Component analysis}
\textbf{Comparison of different perturbation methods.}
Our proposed method works well in all settings, exhibiting its strong adaptability to various scenarios. Overall, we have 42 settings under the node classiﬁcation task, each of which is a combination of different backbone models, different forms of perturbation and datasets. (e.g., GCN-Random-Cora). It is shown that PerturbEmbedding achieves optimal results in 34 settings. The reason why PerturbEmbedding outperforms other methods is that other methods can be regarded as a special form of PerturbEmbedding, directly applying perturbations to embedding could achieve finer granularity of perturbation, thus it could be adapted to various scenarios, making PerturbEmbedding the most ﬂexible and useful method.

\textbf{Uniformity analysis.} We plot the node Embedding Distribution with Gaussian kernel density estimation (KDE) in R2 and von Mises-Fisher (vMF) KDE on angles. Figure \ref{cora_emb} and \ref{cchameleon_emb} show the node Embedding Distribution of GCN and our method trained on cora and chameleon. {We observe that the features learned from PerturbEmbedding (right in Figure \ref{cora_emb} and \ref{cchameleon_emb}) are more uniform compared with GCN (left in Figure \ref{cora_emb} and \ref{cchameleon_emb}), which explains PerturbEmbedding could preserve more information of the data than GCN \cite{wang2020understanding}, thus improves the robustness and generalization abilities of GNNs.
}

\subsection{Efficiency analysis}

\textbf{Training process analysis.} We conduct experiments to analyze the loss during the training process when employing different adversarial methods. Figure \ref{training_loss} shows the change of loss in GCN training processes when employing different adversarial methods on Cora. Furthermore, similar training loss curves can be drawn under other experimental settings. The experimental results suggest that PerturbEmbedding presents the smallest sample variance among all methods, thus achieving the fastest convergence and the most stable performance.

\textbf{Analysis on Computational Efficiency}. We compare the training time of different perturbation methods on the Cora dataset, using GCN as the backbone model. In Table \ref{tab:time}, we observe that PerturbEmbedding is faster than other perturbation methods, and has the closest time to the GCN.

\begin{minipage}{.5\columnwidth}
\centering
\small
\captionof{table}{Averaged running time of 50 epochs on Cora.}
\label{tab:time}
\scalebox{0.75}{
\renewcommand{\arraystretch}{0.9}
\begin{tabular}{cccccccc}
\noalign{\smallskip}\noalign{\smallskip}\hline
        \toprule[1.5pt]
Method & Time (s)  \\
   \toprule[1.0pt]
GCN &0.275 &   \\    
GCN + PerturbNode &  0.362 &\\
GCN + PerturbEdge& 0.335 &\\ 
GCN + PerturbWeight & 0.321 &\\
\textbf{GCN + PerturbEmbedding} & 0.280 &\\
        \toprule[1.5pt]    
    \end{tabular}}

\end{minipage}
\hfill%
\begin{minipage}{.4\linewidth}
\vspace{-2ex}
\begin{figure}[H]
\centering
\includegraphics[width=1.1\linewidth]{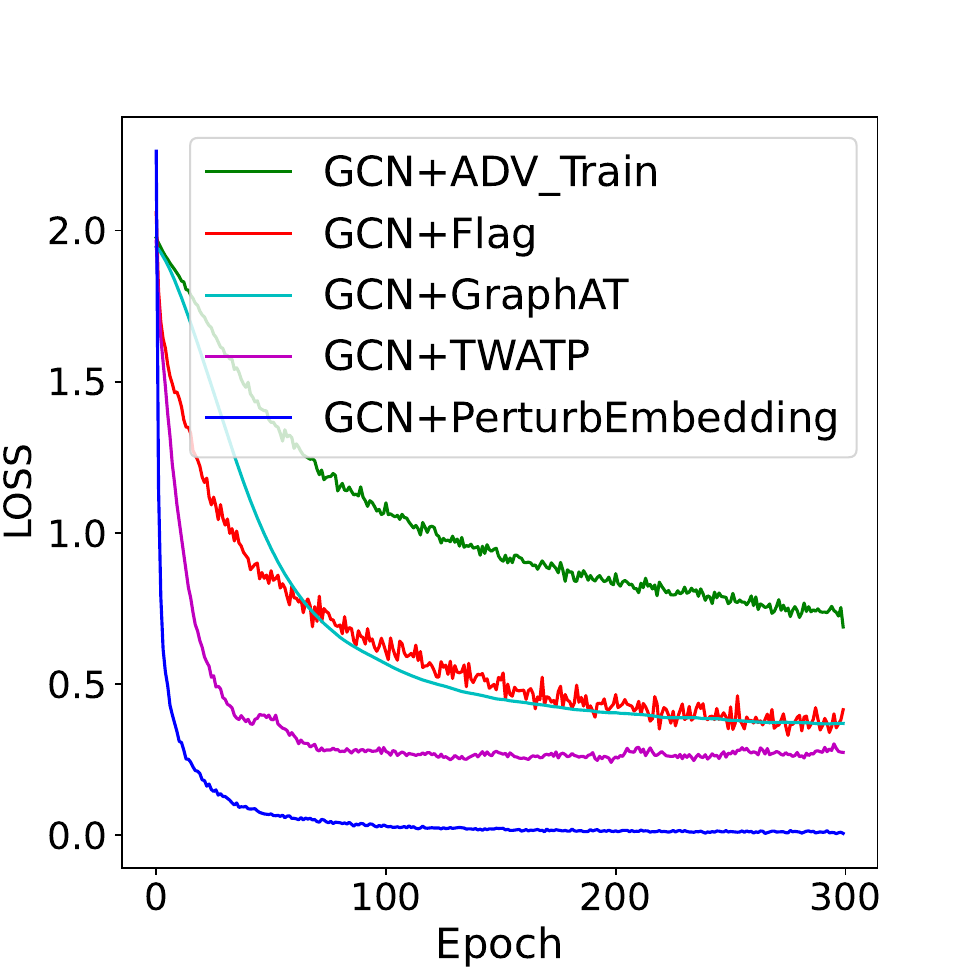}
\end{figure}
\vspace{-5ex}  
\captionof{figure}{Training loss.}
\vspace{-2ex}  
\label{training_loss}

\end{minipage}

\section{Conclusion}
 
In this paper, we propose a novel method, PerturbEmbedding, that integrates adversarial perturbation and training, enhancing GNNs' resilience to such attacks and improving their generalization ability. PerturbEmbedding performs perturbation operations directly on every hidden embedding of GNNs and provides a unified framework for most existing perturbation strategies/methods. We also offer a unified perspective on the forms of perturbations, namely random and adversarial perturbations. 
Through experiments on various datasets using different backbone models, we demonstrate that PerturbEmbedding significantly improves both the robustness and generalization abilities of GNNs, outperforming existing methods. The rejection of both random (non-targeted) and adversarial (targeted) perturbations further enhances the backbone model's performance. 

\clearpage
\bibliographystyle{ACM-Reference-Format}
\bibliography{file}
\clearpage
\begin{table*}[]
\captionsetup{font={small,stretch=1.25}, labelfont={bf}}
 \renewcommand{\arraystretch}{1}
    \caption{Statistics of the graph datasets. \#C is the number of distinct node classes, h is the homophily ratio.}
    \label{tab:datasets statistics}
    \centering
        \resizebox{0.99\linewidth}{!}{
    \begin{tabular}{cccccccccc}
        \toprule[1.5pt]
        \textbf{Dataset}      & \textbf{Nodes}& \textbf{Edges} & \textbf{Features} & \textbf{\#C}&  \textbf{Edge hom}& \textbf{h} & \textbf{\#Training Nodes} & \textbf{\#Validation Nodes}&  \textbf{\#Testing Nodes}  \\        \toprule[1.0pt]
	 Penn94 & 41,554 & 1,362,229 & 5 & 2 & .470 & .046 & 50\% of nodes per class   & 25\% of nodes per class & Rest nodes\\
  
	 chameleon &  2,277 & 3,6101  & 2,325 & 5 & .23 & .062 & 48\% of nodes per class   & 32\% of nodes per class & Rest nodes\\
  
	 film &  7,600 & 29,926 & 931 & 5  & .22 & .011 & 48\% of nodes per class   & 32\% of nodes per class & Rest nodes\\
  
      squirrel &  5,201 & 216,933  & 2,089 & 5 & .22 & .025 & 48\% of nodes per class   & 32\% of nodes per class & Rest nodes\\
      
      Cora & 2,708 & 5,278 & 1,433 & 7 & .81& .766 & 48\% of nodes per class   & 32\% of nodes per class & Rest nodes \\

      Citeseer & 3,327 & 4,676 & 3,703 & 6 &.74 &.627 & 48\% of nodes per class   & 32\% of nodes per class & Rest nodes \\

      Pubmed & 19,717 & 44,327 & 500 & 3 & .80& .664& 48\% of nodes per class   & 32\% of nodes per class & Rest nodes \\
      
        \toprule[1.5pt]    
    \end{tabular}}
\end{table*}

\appendix

This is the Appendix for ``Unifying Adversarial Perturbation for Graph Neural Networks''.

\section{Preliminary}

\textbf{Message-passing GNNs.} Based on the strong homophily assumption that nodes with similar properties are more likely to be linked together, most of the existing GNN models (GCN, GAT) adopt the message-passing framework, where each node sends messages to its neighbors and simultaneously receives messages from its neighbors. Formally, the $k$-th
iteration of message passing, or the $k$-th layer of GNN forward computation is: 
\begin{align}
    \mathbf{h}^{(k)}_{v} = \operatorname{COMBINE}^{(k)}\left(\mathbf{h}^{(k - 1)}_{v}, \operatorname{AGGR}\left\{\mathbf{h}^{(k - 1)}_{u}: u \in {N}{(v)}
     \right\}\right)
\end{align} 
where $\mathbf{h}^{(k)}_{v}$ is the embedding of node $v$ at the $k$-th layer, and $N(v)$ is a set of nodes adjacent to node
$v$, and $\mathbf{h}^{(0)}_{v} = x_v$. $\operatorname{COMBINE}^{(k)}$ is differentiable function and $\operatorname{AGGR}$ represents the aggregation operation.

A special form of Message-passing GNNs is graph convolutional networks
(GCN) \cite{kipf2016semi}, the forward inference at the $k$-th layer of
GCN is formally defined as:
\begin{align}
    \mathbf{h}^{(k)}_{v} = \sigma\left(\sum_{u \in N(v)}\left(\mathbf{W}^{(k-1)}h_{v}^{(k-1)}\tilde{A}_{uv}\right)\right)
\end{align}
where $\sigma(\cdot)$ is the ReLU function, ${\mathbf{W}^{(k-1)}}$ is the trainable weight matrix at layer k - 1. Let $\tilde{A}_{i:}$ denote the $i$th row of $\tilde{A}$ and $\mathbf{H}^{(k)} = \left[ 
 {(\mathbf{h}^{(k)}_{1})}^T;...; {(\mathbf{h}^{(k)}_{N})}^T\right]$, we than have the standard form of GCN:
 \begin{align}
    \mathbf{H}^{(k)} = \sigma\left(\tilde{\mathbf{A}}\mathbf{H}^{(k-1)} {(\mathbf{W}^{(k-1)})}^T  \right) 
 \end{align}
where  $\mathbf{\hat{A}} = \mathbf{A} + \mathbf{I}, \tilde{\mathbf{A}} = {\mathbf{\hat{D}}}^{-\frac{1}{2}}\mathbf{\hat{A}} {\mathbf{\hat{D}}}^{-\frac{1}{2}}$.

\textbf{LINKX.} However, such an assumption of homophily is not always true for heterophilic graphs, Message-passing GNNs fail to generalize to this setting. \cite{lim2021large} proposes a simple method LINKX that combines
two simple baselines MLP and LINK, which overcomes the scalable issues and achieves superior performance on large graphs. LINKX separately embeds node features and adjacency information with MLPs, combines the embeddings by concatenation, then uses a final MLP to generate predictions:
\begin{align}
    Y = \operatorname{MLP}_f\left(\sigma\left(   W[h_A;h_X]+h_A+h_X      \right)   \right)
\end{align}
where $h_A = \operatorname{MLP}_A(A) $, $h_X = \operatorname{MLP}_X(X)$.

\textbf{Adversarial Training.} Standard adversarial training seeks to solve the min-max problem as:
\begin{align}
    % \min_{\theta} \mathbb{E}_{(x,y) \sim \mathcal{D}} \left[ \max_{\lVert  \delta \rVert_p \le \boldsymbol{\epsilon} } L\left(f_{\theta}\left(x + \delta\right), y \right)\right]
    \min_{\theta} \max_{\lVert  \delta \rVert_p \le \boldsymbol{\epsilon} } L\left(f_{\theta}\left(x + \delta\right), y \right)    
\end{align}

where $y$ is the label, $\lVert  \cdot \rVert$ is some $l_{p}$-norm distance metric, $\boldsymbol{\epsilon}$ is the perturbation budget, and $L$ is the objective function. The typical approximation of the inner maximization under an  $l_{\infty}$-norm constraint is as follows \cite{}:
\begin{align}
    \delta_{t+1} = \begin{matrix} \prod_{\lVert \delta \rVert_{\infty} \le \boldsymbol{\epsilon} } \end{matrix} \left( \delta_{t} + \alpha \cdot sign \left( \nabla_\delta L\left(f_{\theta}\left(x + \delta_{t}\right), y \right)\right)\right) 
\end{align}
where perturbation $\delta$ is updated iteratively, and $\begin{matrix} \prod_{\lVert \delta \rVert_{\infty} \le \boldsymbol{\epsilon} } \end{matrix}$ performs projection onto the $\boldsymbol{\epsilon}$-ball in the $l_{\infty}$-norm. For maximum robustness, this iterative updating procedure usually loops M times, which makes PGD computationally expensive. While there are M forward and backward steps within the process, $\theta$ gets updated just once using the final $\delta_M$.

\section{Algorithm details}
\label{alg:Algorithm_details}

\begin{algorithm}[]
	\caption{Unified adversarial perturbation (targeted) for GNNs}\label{alg:pseudo_code}
	\begin{algorithmic}[1]
			\STATE \textbf{Input:} Graph data $G=(V, E)$ and label $Y$.
			% \State Randomly initialize $\theta$, $\omega_e$ and $\omega_g$.
			\WHILE {Not converged or maximum epochs not reached} 
                    \FOR{\ $t=1 \text { to } T$ }
                        % \STATE Generate targeted adversarial perturbations via three different Generators in Equation \ref{adversarial_unified2}.
                        \STATE Compute loss $L$ in Equation \ref{adversarial_representation}.
                        \IF {t == T}
        			 	\STATE Update the parameters of generator $\beta$ by descending its stochastic gradient.
                        \ELSE
                            \STATE Update the parameters of GNN $\theta$ by descending its stochastic gradient.
                        \ENDIF
    			\ENDFOR 
			\ENDWHILE \\
       The gradient-based updates can use any standard gradient-based learning rule. We used Stochastic gradient descent in our experiments.
	\end{algorithmic}
\end{algorithm}

\section{Dataset Details}
\label{sec:appendix_implementation_detail}

In this paper, we conduct experiments on 7 graph datasets, including three homophilous datasets and 4 heterophilous datasets, here are the details of these 7 graph datasets in our experiments, and dataset statistics are in Appendix Table \ref{tab:datasets statistics}.

\begin{itemize}
    \item \emph{Cora}, \emph{Citeseer} and \emph{PubMed}: These 3 different datasets are most widely used homophilous datasets where majority of edges connect nodes within the same class. Specifically, these 3 datasets are citation graphs, where each node represents a scientific paper. These graphs use bag-of-words representations as the feature vectors of nodes. Each node is assigned a label indicating the research field. 
    \item \emph{Penn94} \cite{traud2012social} is a friendship network from the 2005 Facebook 100 university student networks, where nodes represent students. Each node is labeled with the user's reported gender. Major, second major/minor, dorm/house, year, and high school are the features of the node.
    \item \emph{Film/Actor} is an actor co-occurrence network in which nodes represent actors and edges represent co-occurrence on the same Wikipedia page. It is the actor-only induced subgraph of the film-director-actor-writer network.
    \item \emph{Chameleon and Squirrel} are Wikipedia networks \cite{rozemberczki2021multi}, in which nodes represent Wikipedia web pages and edges are mutual links between pages. And node features correspond to several informative nouns on the Wikipedia pages. Each node is assigned one of five classes based on the average monthly traffic of the web page.
\end{itemize}

\end{document}